\documentclass[UTF8]{article}

\usepackage{xspace}

\usepackage[final]{neurips_2025}

\usepackage[utf8]{inputenc} 
\usepackage[T1]{fontenc}    
\usepackage[pagebackref=true,breaklinks=true,
colorlinks,
]{hyperref}

\usepackage{tablefootnote} 
\usepackage{url}            
\usepackage{booktabs}       
\usepackage{amsfonts}       
\usepackage{nicefrac}       
\usepackage{microtype}      
\usepackage{xcolor}         
\usepackage{longtable}         
\usepackage{makecell}

\usepackage{pifont}
\usepackage{circledsteps}
\usepackage{xltabular}
\usepackage{tocloft}
\usepackage[toc,page,header]{appendix}
\usepackage{adjustbox}
\usepackage{minitoc}
\usepackage{amsthm}

\renewcommand \thepart{}
\renewcommand \partname{}

\usepackage{subcaption}
\usepackage{multicol}
\usepackage{multirow}

\usepackage{pifont}

\usepackage[most]{tcolorbox}
\usepackage{tabularx} 
\usepackage{graphicx} 
\usepackage{geometry} 
\usepackage{enumitem}

\usepackage{wrapfig}
\usepackage{tikz}
\usepackage{comment}
\usepackage{amssymb} 
\usepackage{cleveref} 

\usepackage[table]{xcolor}
\usepackage[dvipsnames]{xcolor}
\usepackage{listings}
\usepackage{caption}
\usepackage{adjustbox}

\usepackage{circledsteps} 
\usepackage{csquotes}
\usepackage{soul} 
\usepackage{color} 
\usepackage{algorithm}
\usepackage{algpseudocode} 
\usepackage{amsmath}        
\usepackage{amssymb}
\usepackage{epigraph}
\newtheorem{theorem}{Theorem}
\usepackage{cases}  
\usepackage{bm} 
\usepackage[draft]{minted}

\newcommand{\ie}{i.e.\xspace}

\newcommand{\rowgray}{\rowcolor{lightgray!30}~}

\newcommand{\inlineColorbox}[2]{\begingroup\setlength{\fboxsep}{1pt}\colorbox{#1}{\hspace*{2pt}\vphantom{Ay}#2\hspace*{2pt}}\endgroup}

\newcommand{\methodshort}{VADTree}

\newcommand{\articletitle}{\methodshort: Explainable Training-Free Video Anomaly Detection via Hierarchical Granularity-Aware Tree}

\title{\articletitle}

\author{%
  Wenlong Li$^{1}$ 
  \quad Yifei Xu$^{1,4}$
  \thanks{Corresponding Author} 
  \quad Yuan Rao$^{1}$ 
  \quad Zhenhua Wang$^{2}$ 
  \quad  Shuiguang Deng$^{3}$\\
  $^{1}$School of Software, Xi’an Jiaotong University\\
  $^{2}$China Railway Xi'an Group\\
  $^{3}$College of Computer Science and Technology, Zhejiang University \\
  $^{4}$ Xi'an Jiaotong University Suzhou Institute \\
  \texttt{wenlongli@stu.xjtu.edu.cn}  \quad 
  \texttt{belonxu\_1@xjtu.edu.cn}\quad 
  \\
}

\begin{document}

\maketitle
\doparttoc 
\faketableofcontents

\begin{abstract}
Video anomaly detection (VAD) focuses on identifying anomalies in videos. 
Supervised methods demand substantial in-domain training data and fail to deliver clear explanations for anomalies. 
In contrast, training-free methods leverage the knowledge reserves and language interactivity of large pre-trained models to detect anomalies. 
However, the current fixed-length temporal window sampling approaches struggle to accurately capture anomalies with varying temporal spans.
Therefore, we propose \textbf{\methodshort}~that utilizes a Hierarchical Granularity-aware Tree (HGTree) structure for flexible sampling in VAD. 
\methodshort~leverages the knowledge embedded in a pre-trained Generic Event Boundary Detection (GEBD) model to characterize potential anomaly event boundaries.
Specifically, \methodshort~decomposes the video into generic event nodes based on boundary confidence, and performs adaptive coarse-fine hierarchical structuring and redundancy removal to construct the HGTree.
Then, the multi-dimensional priors are injected into the visual language models (VLMs) to enhance the node-wise anomaly perception, and anomaly reasoning for generic event nodes is achieved via large language models (LLMs).
Finally, an inter-cluster node correlation method is used to integrate the multi-granularity anomaly scores. Extensive experiments on three challenging datasets demonstrate that \methodshort~achieves state-of-the-art performance in training-free settings while drastically reducing the number of sampled video segments. The code will
be available at~\url{https://github.com/wenlongli10/VADTree}.

\end{abstract}

\section{Introduction}\label{Introduction}

\begin{figure}[!t]
\centering
\includegraphics[
width=1\linewidth]{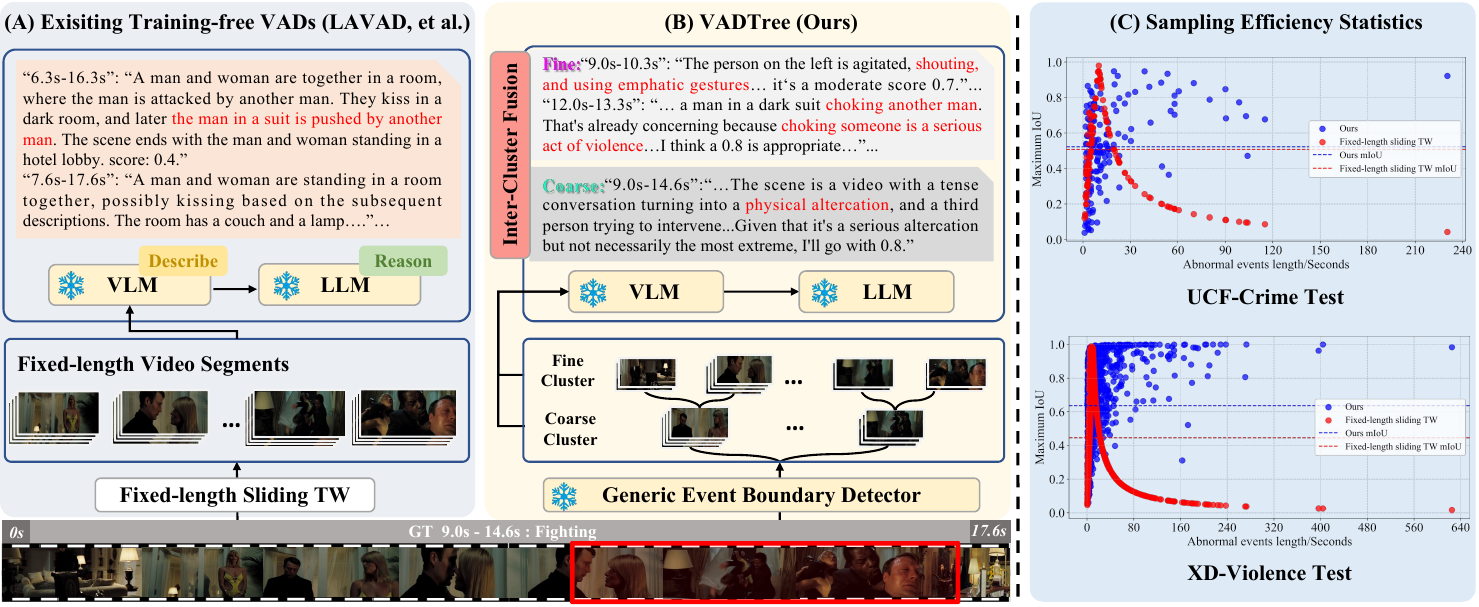}
\caption{
Comparison of our methods with popular paradigms. 
As illustrated in (A), prevailing training-free VAD methods relying on fixed-length sliding temporal window sampling inherently fail to adapt to dynamic anomaly durations. (B) demonstrates our \methodshort~is based on pre-trained knowledge of Generic Event Boundary Detection to achieve adaptive coarse-fine hierarchical representation of videos, and support multi-granularity anomaly understanding and score fusion. 
(C) displays the maximum IoU between all sampled video segments and ground-truth abnormal events across two VAD datasets.
The sampling results of 10 seconds long fixed-length sliding temporal window (TW)~\cite{ye2024vera,zanella2024harnessing} can only achieve higher IoU with abnormal events that are close in length to itself ($\text{mIoU}=0.51$ on UCF-Crime and $\text{mIoU}=0.44$ on XD-Violence). Our granularity-aware tree demonstrates strong flexibility, and achieves higher IoU for events from 3 seconds to 630 seconds, which is the basis for subsequent understanding and localization of anomalies ($\text{mIoU}=0.52$ on UCF-Crime and $\text{mIoU}=0.64$ on XD-Violence). 
}
\label{fig:comparison}
\end{figure}

Video Anomaly Detection (VAD) aims at temporally locating unexpected and unusual events in videos, thereby facilitating  widespread applications including autonomous driving~\cite{yao2022dota,lu2024scaling} and industrial manufacturing~\cite{roth2022towards}. 
Most traditional VAD approaches primarily locate anomalous frames by learning the normal or abnormal patterns from training samples with either fully-supervised~\cite{bai2019traffic, wang2019anomaly}, weakly-supervised~\cite{sultani2018real, chen2024prompt, yang2024text, yan2023feature, lv2023unbiased} or unsupervised learning~\cite{ye2019anopcn,zhang2024multi,lu2013abnormal}. 

While the aforementioned methods perform competitively on experimental VAD benchmarks, their inherent drawbacks limit the capabilities of interpretability, generalization, and interaction in real-world applications. The rapid development of pre-trained Large Language Models (LLMs) and Visual Language Models (VLMs) facilitates the combination of visual comprehending and language interaction, which are particularly well-suited for explainable VAD in real-world surveillance scenarios. Recent research on explainable VAD generates semantic segments of long-term videos with temporal window strategy and equips VLMs with auxiliary guidance to make interpretable anomaly scoring ~\cite{lv2024video,zanella2024harnessing,ye2024vera}. 
As the pioneering training-free VAD, LAVAD~\cite{zanella2024harnessing} exploits an off-the-shelf VLM to caption each video frame, and enables
LLMs to  aggregate and score scene semantic dynamics over time in each temporal window. 
Inspired by Verbalized Machine Learning (VML), VERA~\cite{ye2024vera} leverages video-level annotation data and verbalized learning to optimize a set of guidance questions to drive the frozen VLMs to make abnormal judgments on semantics within the temporal window, and yields frame-level anomaly scores in a coarse-to-fine manner without parameter modifications. 

Towards explainable VAD in more practical real-world scenarios, there remains a significant gap to comprehend and reason about anomalies with different durations. A key challenge lies in accurately localizing diverse anomalies under a training-free setting. 
The video segments sampled by fixed temporal windows are straightforward to implement but remain far from the ground-truth abnormal event boundaries~\cite{lv2024video,zanella2024harnessing,ye2024vera}. 
More critically, this strategy risks abrupt semantic discontinuities or the conflation of irrelevant semantics, which exacerbates the noise in abnormal semantics and amplifies hallucinations of VLMs. 
Although HolmesVAU~\cite{zhang2024holmesvau} introduces an anomaly-focused temporal sampler to handle the anomalies of varying durations, this approach trained on domain-specific videos is prone to underperform in practical videos recorded in changed domains. 
Another fundamental limitation is the inability to comprehensively understand multi-granular anomalies. While existing explainable VADs excel at detecting transient anomalies, such as traffic accidents or explosions, they often fail to model more complex events like burglaries and arrests.
These complex events require extended contextual reasoning. 
Previous studies~\cite{zanella2024harnessing,ye2024vera} have attempted to integrate semantics from sliding temporal windows.  However, the fixed-length windows inherently conflicts with the dynamic characteristics of event durations in real-world scenarios, struggling to address frame redundancy and inevitable noise.

To address these challenges, we propose \methodshort~, a training-free VAD framework that realizes multi-granularity anomaly reasoning via hierarchical event-aware video understanding. Unlike existing temporal window-based approaches, \methodshort~adaptively organizes video content into a hierarchical granularity-aware tree structure by leveraging pre-trained generic event boundary detectors. 
This tree structure naturally aligns with the temporal dynamics of real-world events, allowing for adaptive sampling of video segments that match anomaly durations. 
We address semantic noise in anomaly scoring by introducing intra-cluster node refinement that aggregates contextually relevant nodes to refine initial predictions. 
To resolve the conflicts between coarse-grained and fine-grained cues, we develop inter-cluster node correlation to dynamically integrate anomaly evidence across temporal granularities, enhancing detection robustness through score consistent aggregation. 
We evaluate \methodshort~on three benchmark datasets: UCF-Crime~\cite{sultani2018real}, XD-Violence~\cite{wu2020not}, and MSAD~\cite{zhu2024advancing}.
Our empirical results demonstrate that \methodshort~outperforms unsupervised, one-class, and training-free VAD methods.
This work makes the following contributions:
\begin{itemize}[itemsep=0pt,topsep=0pt,parsep=0pt]
    \item 
    We propose \textbf{\methodshort}, a training-free generic event-centric video anomaly detection framework that flexibly leverages pre-trained GEBD knowledge to localize anomalous events in temporal positions. VADTree overcomes the inefficiency and roughness of dense sampling while providing a multi-granularity perception and reasoning capability for training-free VAD.
    \item We propose a hierarchical granularity-aware tree that utilizes a coarse-fine representation of anomalous videos based on potential generic event boundaries. Additionally, we design an event-centric anomaly scoring and refining approach to derive generic event anomaly scores from tree nodes, which integrates multidimensional prior information and multi-granularity scores to enhance VAD performance and reasoning ability.
    \item \methodshort~achieves SOTA performance among training-free, unsupervised, and one-class methods on both UCF-Crime and XD-Violence datasets, and even surpasses some weakly supervised methods on MSAD dataset. 
\end{itemize}

\section{Related Work}

\subsection{Video Anomaly Detection}
Traditional VAD approaches primarily employ deep neural networks (DNNs) through three dominant learning paradigms. 
Fully-supervised methods~\cite{bai2019traffic, wang2019anomaly} utilize frame-level annotations to learn the distinction between normal and abnormal frames, but they entail a prohibitive cost of acquiring large-scale labeled datasets. 
Weakly-supervised approaches~\cite{sultani2018real, chen2024prompt, yang2024text, zhang2023exploiting, lv2023unbiased} address this limitation by training discriminative models using video-level labels from both normal and abnormal samples, learning to identify anomalous patterns without precise temporal annotations. 
Unsupervised learning approaches~\cite{liu2018future, ye2019anopcn, zhang2024multi, wang2019gods, lu2013abnormal, tur2023unsupervised} bypass annotation requirements entirely by solving frame reconstruction or prediction tasks to construct distinct representation spaces for normal and anomalous video content. 
The traditional methods still lack interactivity and rely heavily on the availability of training data. 

Recent advances~\cite{zhang2024holmesvau, zhang2024holmesvad, zanella2024harnessing, yang2024follow, lv2024video,ye2024vera,gao2025suvad} have successfully leveraged VLMs to generate interpretable textual descriptions of detected anomalies. Current approaches primarily follow two paradigms: 
(1) Methods that rely on frozen models first split videos via sliding temporal windows, then analyze potential anomalies through multiple pre-trained models~\cite{zanella2024harnessing,ye2024vera,gao2025suvad,dev2025mcanet}. 
(2) The instruction fine-tuning based methods utilize DNN-based VAD models to filter out potential abnormal frames, which are then fed into the VLMs along with prompts for further abnormal description and judgment~\cite{zhang2024holmesvau,zhang2024holmesvad}. 
However, the sliding temporal window employed in current training-free methods suffers from inflexibility and sampling redundancy, making it challenging to accurately capture anomalous events with varying content lengths. Instruction fine-tuning based methods require additional data and computational resources to identify potential anomalous video segments. 
In our work, we explore an adaptive temporal sampling approach for potential anomalous events under training-free conditions.

\subsection{Event-based Video Understanding}
An event is an inherent semantic unit of videos, serving as a critical foundation for scene context understanding. Recent advances in video understanding have extensively explored event-centric representations to achieve compact and effective modeling~\cite{wang2024videotree,guo2024trace,cheng2024enhancing,jiang2024prior}.  Specifically, HEMLLM~\cite{cheng2024enhancing} designs an adaptive sequence segmentation mechanism to partition long videos into coherent event segments. Similarly, LLMEPET~\cite{jiang2024prior} employs pseudo-events to guide precise moment prediction within event boundaries. TRACE~\cite{guo2024trace} introduces a causal event modeling framework to deconstruct videos into event sequences, where the current event is predicted based on previous event information and textual instructions. Meanwhile, VideoTree~\cite{wang2024videotree} constructs a query-adaptive hierarchical representation grounded in the inherent event and scene structure of videos.
As an event-aware  training-free VAD framework, EventVAD~\cite{shao2025eventvad} integrates dynamic spatiotemporal graph modeling and VLMs to detect anomaly events.
However, the generalization of the event-aware method it constructs has not been verified, and its robustness in identifying complex boundaries is limited.
Differently, our work focuses on training-free VAD that addresses anomalous semantic understanding within multi-granularity generic event-structured video representations.

\section{Methodology}
\begin{figure}[!t]
\centering
\includegraphics[width=1\linewidth]{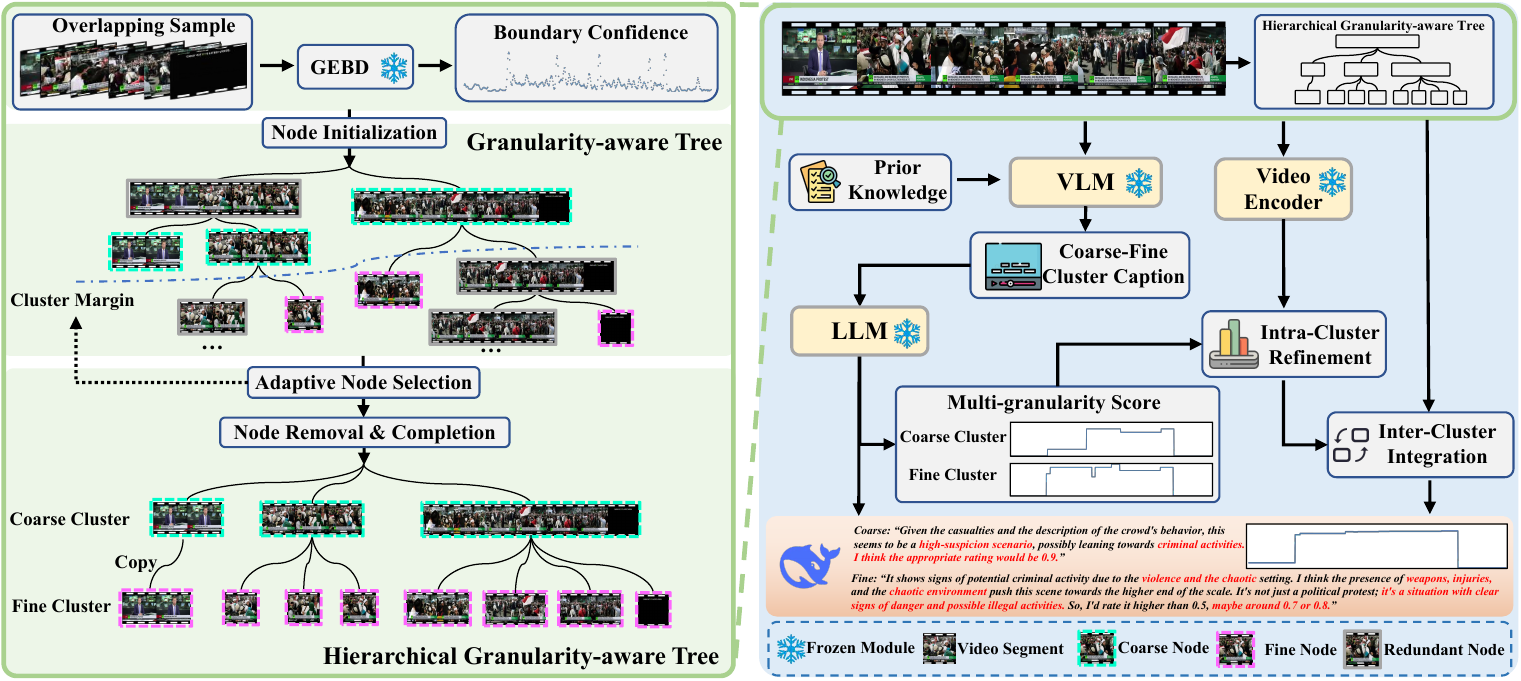}

\caption{The architecture of our proposed~\methodshort. The left side shows the construction of a hierarchical granularity-aware tree, which provides flexible multi-granularity characterization for the understanding and location of abnormal events. Then, as shown on the right, the description, reasoning, and refinement are implemented in a node-wise manner, and finally abnormal score integration is completed based on the topological relationship of the HGTree.}
\label{fig:method}
\end{figure}
Given an input video sequence ${V} = \{I_t\}_{t=1}^T$ with $T$ frames, our training-free approach aims to directly locate and detect the anomalous events within ${V}$ without any parameter updates or fine-tuning on external datasets. 
The overall pipeline of our \methodshort~is illustrated in Figure~\ref{fig:method}, which is composed of a hierarchical granularity-aware tree, generic event-centric anomaly scoring and refining, and inter-cluster node correlation. 
Firstly, we utilize the GEBD pre-trained model and depth-first traversal to construct a granularity-aware tree. Further, we use k-means clustering to stratify and simplify the tree, resulting in a hierarchical granularity-aware tree with coarse and fine clusters (Section~\ref{sec:3.1}).
Then, the generic event-centric anomaly scoring module produces initial anomaly score based on a video content description and intrinsic prior knowledge, and ensures contextual relevance and reduces scoring inaccuracy by aggregating score from semantically similar segments in the intra-cluster.
For the fusion of inter-cluster anomaly scores, we design a cohesion-driven correlation mechanism to ensure semantic integrity and complementarity across different hierarchical structures (Section~\ref{sec:3.2}). 
\subsection{Hierarchical Granularity-aware Tree} \label{sec:3.1}
To address the inherent limitations of uniform sampling for arbitrary-length anomalies, we construct a hierarchical granularity-aware tree by leveraging pre-trained GEBD knowledge~\cite{zheng2024rethinking,shou2021generic}, which adaptively accommodates events with diverse temporal scales through dynamic multi-granularity decomposition. 
It mainly includes three operations: segmentation confidence sequence generation, generic event node initialization, and adaptive node stratification. 
\paragraph{Segmentation Confidence Sequence} \label{sec:3.1.1}
As conventional GEBD models are limited to processing short video clips (duration of $l_{raw}$ frames), we extend their capability to long videos through an overlapping sliding window strategy. Inspired by~\cite{zhu2023autoshot,soucek2024transnet}, we first partition the input video $V$ into  \( K \) overlapping temporal segments \( \{V_{local}^{(k)}\}_{k=1}^K \). 
Each segment is independently processed by the pre-trained GEBD model to generate a local boundary confidence sequence $C_{local}^{(k)}$ of length $l_{raw}$:   
\begin{equation}
C_{local}^{(k)} = \left[ (t_1, c_1), (t_2, c_2), \dots, (t_{l_{raw}}, c_{l_{raw}}) \right] = f_{\text{GEBD}}\left( V_{local}^{(k)} \right),
\label{eq:gebd}
\end{equation}
where \( t_i \in \mathbb{Z}^+ \) represents the global frame index in $V$, and \( c_i \in [0,1] \) is the confidence score at position \( t_i \). To mitigate edge effects from windowing, we retain only the central \( l_{raw}/2 \) frames from each \( C_{local}^{(k)} \).
 These partial sequences are concatenated into a unified global confidence sequence $C$ that preserves both positional information and confidence scores:
\begin{equation}
\label{eq:peak}
    \begin{aligned}
    {C} &= \operatorname{Concat}\limits_{k=1}^K \left( C_{local}^{(k)}\left[ 
\lfloor \tfrac{1}{4}l_{raw}\rfloor:\lfloor\tfrac{3}{4}l_{raw}\rfloor \right] \right), \\
    \hat{{C}} &= \{ (t, \hat{c}) \, \big| \, \operatorname{LocalMax}({C}, t )\},
    \end{aligned}
\end{equation}
where \(\operatorname{Concat(\cdot)}\) operator aligns confidence scores by their global indices \( t \), and $\lfloor \cdot \rfloor$ ensures the central
frames is an integer. 
This operation essentially discards the fractional part of the division result. $\operatorname{LocalMax(\cdot)}$ identifies peak positions at index $t$ if the confidence score $C(t)$ satisfies:
\begin{equation}
C(t) \geq C(t \pm 1).
\end{equation}

\paragraph{Generic Event Node Initialization}
\label{sec:3.1.2}
\methodshort~constructs a granularity-aware binary tree $\mathcal{T} = \{ \mathcal{N}_i \}_{i=0}^M$ where each node $\mathcal{N}_i = (\hat{c}_l^{(i)}, \hat{c}_r^{(i)}, V_{l:r}^{(i)})$ represents a video segment $V_{l:r}^{(i)}$ with associated confidence scores $\hat{c}_l^{(i)}$ and $\hat{c}_r^{(i)}$ for its left and right temporal boundaries.
The tree structure is built via the $\operatorname{TreeInit}$ algorithm (Appendix~\ref{app:hgtree}), which recursively splits segments at the most confident event boundaries $\hat{c}_{\text{max}} \in \hat{C}$.
Initialization starts with the root node $\mathcal{N}_0 = (1, 1, V_{1:T}^{(0)})$, where $1$ means that the confidence of the factual boundary for the beginning and end frames. The algorithm performs depth-first partitioning until  either exhausting all candidate boundaries or encountering confidence values below the threshold $\gamma_{\text{min}}$. The resulting tree $\mathcal{T}$ inherently encodes temporal granularity through its hierarchical organization, where internal nodes represent segmentation decisions and leaf nodes correspond to atomic events.

\paragraph{Adaptive Node Stratification}\label{sec:3.1.3}
After initializing the granularity-aware tree $\mathcal{T}$, we stratify it hierarchically to enable multi-granular representation of videos.  
Given the continuous confidence scores and the inherent content uncertainty in videos, 
$\mathcal{T}$ allows for decomposition into an arbitrary number of hierarchical clusters. Considering the marginal performance gains diminishing with excessive layers and the resultant increase in inference overhead, 
we adopt the classic two cluster granularity semantic perception strategy to partition $\mathcal{T}$ into coarse cluster parent-wise nodes and fine-cluster child-wise nodes~\cite{xu2024slowfast}. The former captures those clear event boundaries, while the latter captures localized motion patterns over shorter temporal intervals. To account for varying boundary clarity across scenarios and filming conditions, we dynamically determine these clusters via K-means clustering, thereby evolving $\mathcal{T}$ into a hierarchical granularity-aware tree $\mathcal{T}^{'}=\{\mathcal{S}^{'}_{coarse}, \mathcal{S}^{'}_{fine}\}$.
\begin{equation}
\begin{aligned}
& (\hat{C}_{coarse}, \hat{C}_{fine}) = \operatorname{K-Means}(\hat{C}, 2), \\
& \mathcal{S}_{coarse} = \left\{ \mathcal{N}_i \,\big|\, \min({\hat{c}_l}^i, {\hat{c}_r}^i) \geq \min(\hat{C}_{coarse}) \right\}, \quad
\mathcal{S}_{fine} = \left\{ \mathcal{N}_i \,\big|\, \min({\hat{c}_l}^i, {\hat{c}_r}^i) \leq \max(\hat{C}_{fine}) \right\}, \\
& \mathcal{S}^{'}_{coarse} = \operatorname{RemoveDup}(\mathcal{S}_{coarse}), \quad
\mathcal{S}^{'}_{fine} = \operatorname{Complete}(\operatorname{RemoveDup}(\mathcal{S}_{fine})),
\end{aligned}
\label{eq:partition}
\end{equation}
where $\operatorname{RemoveDup}(\cdot)$ and $\operatorname{Complete}(\cdot)$ denote the redundancy elimination and node completion operators, respectively. 
These two clusters $\mathcal{S}^{'}_{coarse}$ and $\mathcal{S}^{'}_{fine}$ are constructed per Eq.~\ref{eq:partition}, by applying K-means clustering and the comparison operation to the confidence scores $\hat{c}$.

To achieve maximal granularity,  $\operatorname{RemoveDup}(\cdot)$ is applied to retain only the finest-grained nodes along each tree path while pruning redundant ancestor nodes. Besides, as some leaf nodes in 
$\mathcal{S}^{'}_{coarse}$ cannot be further split and therefore lack corresponding fine-grained segments,  $\operatorname{Complete}(\cdot)$ function replicates these critical nodes to ensure comprehensive coverage. Crucially, the nodes in both $\mathcal{S}^{'}_{coarse}$ and $\mathcal{S}^{'}_{fine}$ can guarantee complete video representation. 
The details of  $\operatorname{RemoveDup}(\cdot)$ and $\operatorname{Complete}(\cdot)$  along with the proof of comprehensive coverage are provided in~\ref{app:proof}.

\subsection{Generic Event-centric Anomaly Scoring and Refining}\label{sec:3.2}

\paragraph{Prior-infused Node Scoring}
When humans recognize behavior, well-learned societal scripts inherently trigger cognitive associations~\cite{schank1977scripts}. Building on this foundation, we consider these observed patterns as priors for anomaly analysis, which can be systematically categorized along three dimensions: event scene $b_{scene}$, specific characters/objects $b_{obj}$, and actions/behaviors $b_{act}$. Particularly, we explicitly exclude two categories of ill-posed semantic cues for VLMs: (1) \textit{micro-expressions} (e.g., distracted gaze indicating theft intention) that demand prohibitively high image resolution, and (2) \textit{audio-dependent semantic triggers} (e.g., loud sounds suggesting explosions) that are unavailable in visual-only surveillance systems. According to the above findings, the LLM processes both generation instructions $P_b$ and constraint instructions $P_c$ to derive multidimensional priors as Eq.~\ref{eq:prior}:
\begin{equation}
\label{eq:prior}
B = (b_{scene}, b_{obj}, b_{act}) = f_{\text{ LLM}_{gen}}(P_b \circ P_c).
\end{equation}
These priors are then injected into VLMs to facilitate human-like reasoning during video content description.  For the HGTree $\mathcal{T}^{'}=\{\mathcal{S}^{'}_{coarse}, \mathcal{S}^{'}_{fine}\}$, 
let $V^{g}_u$ represent the sampled frames at node $u$ in $\mathcal{S}^{'}_g$. The VLM generates content captions via Eq.~\ref{eq:description}:
\begin{equation}
d^{g}_{u} = f_{\text{VLM}}(V^{g}_u, B \circ P_d).
\label{eq:description}
\end{equation}
Following LAVAD~\cite{zanella2024harnessing}, we instruct an LLM to quantify anomaly likelihood through discrete scoring $a \in \{0,0.1,...,1\}$, with 0 and 1 encoding normal and anomalous extremes respectively. The score derivation from prompt $P_s$ follows:
\begin{equation}\label{eq:scoring}
	a^{g}_{u} = f_{\text{ LLM}}(d^{g}_{u}, {P}_s).
\end{equation}

\paragraph{Intra-cluster Node Refinement}
The score derived from Eq.~\ref{eq:scoring} only examines a partial interval in the entire video without considering long-term context, which is prone to local false positive anomalies caused by mutations. To alleviate this issue, we refine the initial anomaly score by taking into account the context of intra-cluster event nodes. Obviously, within the same cluster, nodes with high semantic similarity should logically exhibit converging anomaly scores.
To quantify semantic similarity between different nodes, we  compute cosine similarity 
$\operatorname{sim(\cdot,\cdot)}$ based on their feature representations extracted from a pre-trained vision encoder~\cite{girdhar2023imagebind}. 
For the $V^{g}_u$ of node $u$ in $\mathcal{S}^{'}_g$, let $\kappa_{u} = [\kappa_{u}^{(1)}, \dots, \kappa_{u}^{(K)}]$ index the top-$K$ most similar nodes. 
As shown in Eq.~\ref{eq:refine}, the refined anomaly score $\hat{a}_{u}$ is computed as an ensemble of initial scores 
of top-$K$ nodes relevant to $V^{g}_{u}$. 
\begin{equation}\label{eq:refine}    
    \hat{a}^{g}_u = \underbrace{\sum_{i=1}^K a_{\kappa_u^{(i)}}}_{\text{Initial scores}} 
    \cdot 
    \underbrace{\frac{\exp(\operatorname{sim}(u,\kappa_u^{(i)})/\tau)}{\sum_j^K \exp(\operatorname{sim}(u,\kappa_u^{(j)})/\tau)}}_{\text{Softmax weights}}
\end{equation}

\paragraph{Inter-cluster Node Correlation}
Previous VAD studies~\cite{zhang2024multi,zhang2024holmesvau} have demonstrated the significance of multi-scale learning, given the varying temporal durations of anomalies and the influence of contextual lengths on anomaly determination. 
This observation aligns with our experimental findings that abnormal event cues exhibit cluster-specific variations.
Therefore, we elucidate a cohesion-driven fusion mechanism for multi-granularity decision fusion based on coarse and fine cluster nodes in the hierarchical granularity-aware tree $\mathcal{T}^{'}$. To integrate multi-granularity anomaly cues while suppressing hierarchical inconsistencies, this mechanism dynamically weights the contributions of parent and child nodes through intra-cluster cohesion metrics.

Specifically,
parent nodes and child nodes come from the coarse and fine clusters of the event tree, respectively. 
For each parent node $\mathcal{N}_i$ containing $m$ child nodes $\{\mathcal{N}_{i1}, ..., \mathcal{N}_{im}\}$, we compute the intra-cluster cohesion $w_i$ as the variance of their denoised anomaly scores via Eq.~\ref{eq:var}:
\begin{equation}
	w_i = \frac{1}{m} \sum_{j=1}^m \left( \hat{a}_{{n}_{ij}} - \mu_i \right)^2, \quad \text{where } \mu_i = \frac{1}{m} \sum_{j=1}^m \hat{a}_{{n}_{ij}}.
\label{eq:var}
\end{equation}
Subsequently, we conduct normalization within the coarse clustering process to obtain the $\hat{w}_i$. A lower $\hat{w}_i$ signifies strong semantic consistency among child nodes, implying that the parent node should dominate the fusion process. Conversely, higher $\hat{w}_i$ implies conflicting child node evidence, and the parent node may have missed some instantaneous cue, necessitating greater reliance on the child node's fine semantics. 

By adjusting the initial fusion weight of the 0.5 through the control coefficient $\beta \in [-1,1]$, the final frame-wise anomaly score $\bar{a}$ for each segment is determined based on the anomaly scores of final fine cluster nodes $\bar{a}_{n_{ij}}$:
\begin{equation}
\bar{a}_{n_{ij}} = \frac{1}{2}(1 - \beta \hat{w}_i)\hat{a}_{n_i} + \frac{1}{2}(1 + \beta \hat{w}_i) \hat{a}_{n_{ij}}.
\label{eq:final score}
\end{equation}

\section{Experiments}\label{sec:exp}

\begin{table}[tbp]
    \parbox[t]{.46\linewidth}{
    \caption{
    Results on UCF-Crime dataset show that \methodshort~substantially outperforms all \inlineColorbox{Gray!30}{Training-free}, one-class, and unsupervised methods, even surpassing some weakly-supervised approaches.
        }
     \label{tab:ucf_results}
    \centering 
    \resizebox{\linewidth}{!}{\begin{tabular}{lcc}
                \toprule
                \textbf{Method} & \textbf{Supervision} & \textbf{AUC (\%)} \\
                \midrule
                \multicolumn{3}{c}{\textit {Non-Explainable VAD Methods}}\\
                Sultani \textit{et al.}~\cite{sultani2018real}  &Weakly Supervised& 75.41\\
                Sultani \textit{et al.}~\cite{sultani2018real}  &Weakly Supervised& 77.92\\
                IBL~\cite{zhang2019temporal} &Weakly Supervised & 78.66\\
                GCL~\cite{zaheer2022generative} &Weakly Supervised & 79.84\\
                GCN~\cite{zhong2019graph} &Weakly Supervised & 82.12\\
                MIST~\cite{feng2021mist}  &Weakly Supervised& 82.30\\
                Wu \textit{et al.}~\cite{wu2020not} &Weakly Supervised & 82.44\\
                CLAWS~\cite{zaheer2020claws} &Weakly Supervised & 83.03\\
                RTFM~\cite{tian2021weakly}  &Weakly Supervised& 83.31\\
                RTFM~\cite{tian2021weakly}  &Weakly Supervised& 84.03\\
                Wu $\&$ Liu \textit{et al.}~\cite{wu2021learning}  &Weakly Supervised& 84.89\\
                MSL~\cite{li2022self}  &Weakly Supervised& 85.30\\
                MSL~\cite{li2022self}  &Weakly Supervised& 85.62\\
                S3R~\cite{wu2022self}  &Weakly Supervised& 85.99\\
                MGFN~\cite{chen2023mgfn}  &Weakly Supervised& 86.67\\
                MGFN~\cite{chen2023mgfn} &Weakly Supervised & 86.98\\
                SSRL~\cite{li2022scale}  &Weakly Supervised& 87.43\\
                CLIP-TSA~\cite{joo2023clip} &Weakly Supervised & 87.58 \\
                GS-MoE~\cite{d2025mixture} &Weakly Supervised & 91.58 \\
                UR-DMU~\cite{zhou2023dual} &Weakly Supervised & 86.97\\
                UMIL~\cite{lv2023unbiased} &Weakly Supervised & 86.75\\
                $\pi$-VAD~\cite{majhi2025just} &Weakly Supervised & 90.33\\
                SVM~\cite{sultani2018real} & One Class& 50.00\\
                SSV~\cite{sohrab2018subspace}  &One Class& 58.50\\
                BODS~\cite{wang2019gods}  & One Class&68.26\\
                GODS~\cite{wang2019gods}  &One Class& 70.46\\
                GCL~\cite{zaheer2022generative}  &Unsupervised& 74.20\\
                Tur~\cite{tur2023unsupervised}   &Unsupervised& 66.85\\
                DyAnNet~\cite{thakare2023dyannet}  &Unsupervised& 79.76\\
                \midrule
                \multicolumn{3}{c}{\textit {Explainable VAD Methods}} \\
                VADor~\cite{lv2024video}&Fine-tuning&88.13\\
                Holmes-VAD~\cite{zhang2024holmesvad}&Fine-tuning&89.51\\
                Holmes-VAU~\cite{zhang2024holmesvau}&Fine-tuning&88.96\\
                VERA~\cite{ye2024vera}&Verbalized Learning&86.55\\
               \rowgray  Blip2~\cite{li2023blip}& Training-free&46.42\\
               \rowgray  ZS CLIP~\cite{radford2021learning}  & Training-free & 53.16\\
                \rowgray  ZS ImageBind (Image)~\cite{girdhar2023imagebind} & Training-free & 53.65\\
               \rowgray  ZS ImageBind (Video)~\cite{girdhar2023imagebind} & Training-free & 55.78\\
               \rowgray  LLaVA-1.5~\cite{liu2023improved} & Training-free & 72.84\\
               \rowgray  Video-Llama2~\cite{zhang2023video}& Training-free & 74.42\\
               \rowgray  LAVAD~\cite{zanella2024harnessing} & Training-free & 80.28\\
               \rowgray  SUVAD~\cite{gao2025suvad} & Training-free & 83.90\\
               \rowgray  MCANet ~\cite{dev2025mcanet} & Training-free & 82.47\\
               \rowgray  EventVAD~\cite{shao2025eventvad} & Training-free & 82.03\\
               \rowgray  \textbf{\methodshort (Ours)} & Training-free & \textbf{84.74}\\
                \bottomrule
                \end{tabular}%
    }
}
\hfill
    \parbox[t]{.52\linewidth}{
    \caption{
         Results on XD-Violence dataset demonstrate that \methodshort~achieves significantly superior performance over current \inlineColorbox{Gray!30}{Training-free} approaches in terms of AUC ROC, while also outperforming all one-class and unsupervised methods. The best results among training-free methods are highlighted in bold.
         * denotes the method that incorporates an additional audio modality. \methodshort* employs Kimi-Audio-7B-Instruct to extract audio captions and enables the LLM to perform anomaly reasoning based on the multimodal text information.
         }
         \label{tab:xd_results}
    \resizebox{\linewidth}{!}{\begin{tabular}{lcccc}
            \toprule
             \textbf{Method} &\textbf{Supervision} & \textbf{AP (\%)} & \textbf{AUC (\%)} \\
            \midrule
            \multicolumn{4}{c}{\textit {Non-Explainable VAD Methods}} \\
            Wu \textit{et al.}~\cite{wu2020not}  &Weakly Supervised& 73.20&- \\
            Wu \textit{et al.}*~\cite{wu2020not}  &Weakly Supervised& 78.64&- \\
            MSL~\cite{li2022self}  & Weakly Supervised&75.53&- \\
            Wu and Liu~\cite{wu2021learning} &Weakly Supervised & 75.90&- \\
            RTFM~\cite{tian2021weakly} &Weakly Supervised &77.81&- \\
            RTFM*~\cite{tian2021weakly} &Weakly Supervised &78.54&- \\
            MSL~\cite{li2022self} &Weakly Supervised & 78.28 \\
            MSL~\cite{li2022self} &Weakly Supervised & 78.58&- \\
            S3R~\cite{wu2022self}  &Weakly Supervised& 80.26&- \\
            MGFN~\cite{chen2023mgfn}  &Weakly Supervised& 79.19&- \\
            MGFN~\cite{chen2023mgfn}  &Weakly Supervised&80.11&- \\
            CLIP-TSA~\cite{joo2023clip} &Weakly Supervised & 82.19&- \\
            GS-MoE~\cite{d2025mixture} &Weakly Supervised & 82.89&94.52 \\
            MACIL-SD*~\cite{yu2022modality} &Weakly Supervised & 83.40&-\\
            UR-DMU*~\cite{majhi2025just} &Weakly Supervised & 81.77&-\\
            $\pi$-VAD*~\cite{majhi2025just} &Weakly Supervised & 85.37&-\\
            Hasan \textit{et al.}~\cite{hasan2016learning}  & One Class &-& 50.32\\
            Lu \textit{et al.}~\cite{lu2013abnormal}  & One Class &-& 53.56\\
            BODS~\cite{wang2019gods}  & One Class&- & 57.32\\
            GODS~\cite{wang2019gods}  & One Class&- & 61.56\\
            RareAnom~\cite{thakare2023rareanom}  &Unsupervised &- & 68.33\\
            \midrule
            \multicolumn{4}{c}{\textit {Explainable VAD Methods}} \\
            Holmes-VAD~\cite{zhang2024holmesvad}&Fine-tuning&90.67&-\\
            Holmes-VAU~\cite{zhang2024holmesvau}&Fine-tuning&87.68&-\\
            VERA~\cite{ye2024vera}&Verbalized Learning&70.54&88.26\\
            \rowgray Blip2~\cite{li2023blip}& Training-free&10.89&29.43\\
            \rowgray ZS CLIP~\cite{radford2021learning}   & Training-free& 17.83 & 38.21\\
            \rowgray ZS ImageBind (Image)~\cite{girdhar2023imagebind}   & Training-free& 27.25 & 58.81\\
            \rowgray ZS ImageBind (Video)~\cite{girdhar2023imagebind}  & Training-free & 25.36 & 55.06\\
            \rowgray LLaVA-1.5~\cite{liu2023improved}  & Training-free& 50.26 & 79.62\\
            \rowgray Video-Llama2~\cite{zhang2023video}& Training-free & 53.57 & 80.21\\
            \rowgray LAVAD~\cite{zanella2024harnessing}& Training-free & 62.01 & 85.36\\
            \rowgray SUVAD~\cite{gao2025suvad}& Training-free &\textbf{70.10} & -\\
            \rowgray MCANet*~\cite{dev2025mcanet}& Training-free &69.72 & 87.43\\
            \rowgray EventVAD~\cite{shao2025eventvad} & Training-free & 64.04 & 87.51\\
            \rowgray \textbf{\methodshort~(Ours)} & Training-free & 67.82& \textbf{90.44}\\
            \rowgray \textbf{\methodshort*~(Ours)} & Training-free & 68.85& \textbf{90.55}\\
            \bottomrule
            \end{tabular}
    }
    }
\end{table}



We validate the performance of \methodshort~on three datasets against state-of-the-art VAD methods trained with different types of supervision, as well as other training-free baselines.
To verify the necessity of each core module, we conduct systematic ablation studies to demonstrate the rationality and effectiveness of \methodshort’s proposed components.
In the following, we first describe the experimental setup in terms of datasets and performance metrics.
We then present and discuss the results in Section~\ref{sec:exp:comparison}, followed by the ablation studies in Section~\ref{sec:exp:ablation}, and conclude with qualitative experiments in Section~\ref{sec:exp:qualitative}.
For more experimental analysis and qualitative results, please refer to the Appendix~\ref{sec:app_results}.

\begin{table}[tbp]
    \parbox[t]{.61\linewidth}{
    \caption{
    Results on MSAD dataset demonstrate that training-free approach \methodshort~delivers competitive performance against existing state-of-the-art weakly-supervised methods.
        }
     \label{tab:msad_results}
    \centering 
    \resizebox{\linewidth}{!}{\begin{tabular}{lcc ccc}
        \toprule
         \textbf {Method} & \textbf {Supervision} &  \textbf{AUC (\%)} &  \textbf{AUC\textsubscript{a} (\%)} & \textbf{AP (\%)} & \textbf{AP\textsubscript{a} (\%)} \\
        \midrule
         RTFM~\cite{tian2021weakly} & Weakly Supervised & 86.65 & - & - & - \\
         MGFN~\cite{chen2023mgfn} & Weakly Supervised & 84.96 & - & - & - \\
         TEVAD~\cite{chen2023TEVAD} & Weakly Supervised & 86.82 & - & - & - \\
         UR-DMU~\cite{majhi2025just} & Weakly Supervised & 85.78 & 67.95 & 67.35 & 75.30 \\
         GS-MoE~\cite{d2025mixture} & Weakly Supervised & 87.72 & 69.54 & 68.26 & 76.68 \\
         $\pi$-VAD~\cite{majhi2025just} & Weakly Supervised & 88.68 & \textbf{71.25} & 71.26 & \textbf{77.86} \\
         \textbf{VADTree (Ours)} & Training-free & \textbf{89.32} & 67.85 & \textbf{71.41} & 75.49 \\
         \bottomrule
        \end{tabular}
    }
}
\hfill
    \parbox[t]{.36\linewidth}{
    \caption{Results of \methodshort~on UCF-Crime 
    dataset with different HGTree construction configuration.
    }
    \label{tab:tree}
    \resizebox{\linewidth}{!}{\begin{tabular}{ccc c}
                \toprule
                
                $\bm{\gamma _{min}}$ & \textbf{Cluster Tool} & \textbf{Clusters}& \textbf{AUC (\%)} \\

                \midrule
                0.3 & - &Fine & 80.89 \\
                0.4 & - &Fine & 82.81 \\%
                0.5 & - &Fine & 80.85 \\
                0.3 & K-Means &Coarse + Fine & 83.74 \\%
                0.4 & K-Means &Coarse + Fine & 84.74 \\%
                0.5 & K-Means &Coarse + Fine & 82.40 \\%
                0.4 & K-Medoids &Coarse + Fine & \textbf{85.24} \\%
                \bottomrule
                \end{tabular}
    }
    }
\end{table}

\begin{table}[tbp]
    \parbox{.37\linewidth}{
    \caption{            Effect of different 
    components on UCF-Crime dataset.}
    \label{tab:ablation_components}
    \centering 
    \resizebox{\linewidth}{!}{\begin{tabular}{lc}
                    \toprule
                    \textbf {Module} & \textbf {AUC (\%)} \\
                    \midrule
                        HGTree Fine Cluster& 71.57\\
                        + Prior-infused Node Scoring&75.67 \\
                        + Intra-cluster Node Refinement        & 83.05 \\
                        + Inter-cluster Node Correlation  & \textbf{84.74} \\
                    \bottomrule
                    \end{tabular}%
    }
}
\hfill
    \parbox{.60\linewidth}{
    \caption{
    Comparison of performance of \methodshort~under different VLM and LLM on UCF-Crime dataset.
    }
    \label{tab:base_model}
    \centering
    \resizebox{\linewidth}{!}{
            \begin{tabular}{lcc}
            \toprule
            \textbf { VLM } & \textbf { LLM } & \textbf { AUC (\%) } \\
                \midrule
            LLaVA-NeXT-Video-7B  & DeepSeek-R1-Distill-Qwen-14B  & \textbf{84.74} \\
            InternVL2\textunderscore5-8B  & DeepSeek-R1-Distill-Qwen-14B  & 83.74 \\
             LLaVA-NeXT-Video-7B  &t5gemma-9B-2B  & 84.00\\
            InternVL2\textunderscore5-8B  & t5gemma-9B-2B  & 83.56 \\
                \bottomrule
                \end{tabular}
    }
    }
\end{table}


\paragraph{Datasets}
We evaluate our method using three commonly used VAD datasets featuring real-world surveillance scenarios, \ie, UCF-Crime \cite{sultani2018real}, XD-Violence~\cite{wu2020not}, and MSAD~\cite{zhu2024advancing}.
\textbf{UCF-Crime} is a large-scale dataset comprising 1900 long untrimmed real-world surveillance videos with 13 types of anomalies.
The training set consists of 800 normal and 810 anomalous videos, while the test set includes 150 normal and 140 anomalous videos. 
\textbf{XD-Violence} is another large-scale dataset for violence detection, comprising 4754 untrimmed videos with audio signals and weak labels that are collected from both movies and YouTube. XD-Violence captures 6 categories of anomalies and it is divided into a training set of 3954 videos and a test set of 800 videos. We also evaluate \methodshort~on \textbf{MSAD} dataset, which provides a greater diversity of real-world scenarios than existing benchmarks.

\paragraph{Performance Metrics}
We measure the VAD performance using the area under the curve (AUC) of the frame-level receiver operating characteristics (ROC) as it is agnostic to thresholding for the detection task. For XD-Violence dataset, we also report the average precision (AP), which refers to the area under the frame-level precision-recall curve, following the established evaluation protocol in~\cite{wu2020not}.

\paragraph{Implementation Details} \label{sec:details}
We use EfficientGEBD~\cite{zheng2024rethinking} as the model $f_{\text{GEBD}}$ for generic event boundary knowledge acquisition, and the overlapping sampling window length $l_{raw}$ follows the 10s window of Kinetics-GEBD~\cite{shou2021generic}. 
The video description model $f_{\text{VLM}}$ and the anomaly reasoning model $f_{\text{LLM}}$ use LLaVA-Video-7B-Qwen2~\cite{zhang2024llavanext-video} and DeepSeek-R1-Distill-Qwen-14B~\cite{deepseekai2025deepseekr1} respectively. In all experiments, the VLM input is configured to a maximum of 64 frames, with the LLM having the thinking mode turned on by default. Although the “Think” mode of DeepSeek-R1-Distill-Qwen-14B incurs additional inference overhead, we still intentionally retain it because it generates valuable intermediate reasoning steps that significantly enhance anomaly interpretation. The video encoder $f_{\text{VE}}$ is provided by ImageBind~\cite{girdhar2023imagebind}.

\subsection{Comparison with State of the Art}\label{sec:exp:comparison}

Most videos in \textbf{UCF-Crime} dataset have low resolution, with mild semantic changes between events within the same video and few shot transitions.
Table~\ref{tab:ucf_results} demonstrates that \methodshort~achieves a substantial superiority over all training-free methods.
In particular, its performance exceeds LAVAD~\cite{zanella2024harnessing} by 4.5\% and surpasses EventsVAD~\cite{shao2025eventvad} by 2.7\%. 

The videos in \textbf{XD-Violence} dataset are primarily sourced from films and TV shows. 
Consequently, the content is more deliberately composed and contains significantly more frequent shot transitions. 
Table~\ref{tab:xd_results} reveals that \methodshort~achieves an AUC ROC 5.1\% higher than LAVAD~\cite{zanella2024harnessing}. And it shows a 2.9\% gain over EventsVAD~\cite{shao2025eventvad}, thereby establishing a new state-of-the-art.
Furthermore, \methodshort~exhibits superior anomaly detection performance compared to all single-class and unsupervised methods on UCF-Crime and XD-Violence datasets. 

Surprisingly, VADTree achieves state-of-the-art performance on \textbf{MSAD} dataset (as shown in Table~\ref{tab:msad_results}) and attaining the highest scores in overall metrics.
This demonstrates that our training-free approach outperforms even weakly-supervised methods that rely on extensive training data. While $\pi$-VAD shows a slight advantage on anomaly-specific metrics AUC\textsubscript{a} and AP\textsubscript{a}, likely due to its supervised fine-tuning on anomalous segments,
VADTree's competitive performance without any dataset-specific training highlights its superior generalization capability.

\subsection{Ablation Study}\label{sec:exp:ablation}
In this section, we present the ablation study conducted on UCF-Crime dataset. We first ablate the effectiveness of each proposed component of \methodshort, and then elaborate on the effect of the HGTree on the final detection accuracy under different construction parameters.
Finally, we discuss the impact of different pre-trained model combinations on the performance of \methodshort~.

\paragraph{Effect of Each Proposed Component}
We ablate different modules of our proposed method \methodshort~to prove the effectiveness of the four proposed components, including HGTree fine cluster, prior-infused node scoring, intra-cluster node refinement and inter-cluster node correlation. 
As shown in Table~\ref{tab:ablation_components}, we first use HGTree fine cluster to express the entire video and build a baseline.
When we input the prior knowledge for the prompt of VLM, the AUC ROC of the method is further improved, indicating that the understanding and accurate description of anomalies can benefit from the prior information of anomalies.
If we further refine the initial anomaly scores of each node within intra-cluster, the AUC ROC will be significantly improved; 
this is because the module can effectively suppress the inference noise and hallucination of VLM and LLM, and introduce references for independent reasoning of each segment. 
Finally, inter-cluster node correlation further increased the AUC ROC to 84.7\%, indicating that the HGTree based parent-child node structure information can effectively guide multi-granularity scores fusion.

\paragraph{Effect of Different HGTree Construction Configurations}
The empirical analysis systematically examines how HGTree's configuration parameters govern VAD performance. 
As illustrated in Table~\ref{tab:tree}, varying the  $\gamma_{min}$ directly modulates the granularity of segmented videos; lower thresholds($\gamma_{min}=0.3$) induce noisy event boundaries due to over-segmentation, whereas higher thresholds($\gamma_{min}=0.5$) restrict hierarchical results diversity. 
Our framework achieves optimal balance at $\gamma_{min}=0.4$, which consistently delivers peak performance across benchmarks. Crucially, ablation studies reveal that using only leaf nodes reduces HGTree to a single cluster structure, yielding a 1.9\% lower AUC ROC compared to the hierarchical two clusters architecture. 
This indicates the importance of hierarchical granularity-aware representation and decision correlation. 
Furthermore, replacing K-Means with K-Medoids for clustering produces an additional performance gain, demonstrating the advantage of using more robust centroid selection when dealing with potential outliers in generic event boundary.

\paragraph{Effect of Different VLM and LLM Configurations}
To comprehensively evaluate the generalizability, we conduct tests with alternative model architectures. Specifically, we select InternVL2\textunderscore5-8B~\cite{chen2024expanding} as an additional VLM with distinct input specifications of 32 frames at 448$\times$448 resolution, contrasting with LLaVA-NeXT-Video-7B's 64 frames, 384$\times$384 inputs. This means that InternVL2\textunderscore5-8B focuses more on perceiving spatial details rather than the temporal continuity of actions.
Furthermore, we incorporate t5gemma-9B-2B~\cite{zhang2025encoder} as additional LLM variant featuring a unique 9B-encoder and 2B-decoder configuration. This differs from DeepSeek-R1-Distill-Qwen-14B’s autoregressive architecture.
As evidenced by the results in Table~\ref{tab:base_model}, the performance fluctuation of \methodshort~is acceptable under different model combinations, which demonstrates the strong generalization ability of our framework.

\subsection{Qualitative Analysis}\label{sec:exp:qualitative}

\definecolor{Mycyan}{HTML}{66FEE2}
\definecolor{Myrose}{HTML}{FF66FE}

\begin{figure}[t] 
\centering 
\includegraphics[width=1\linewidth]{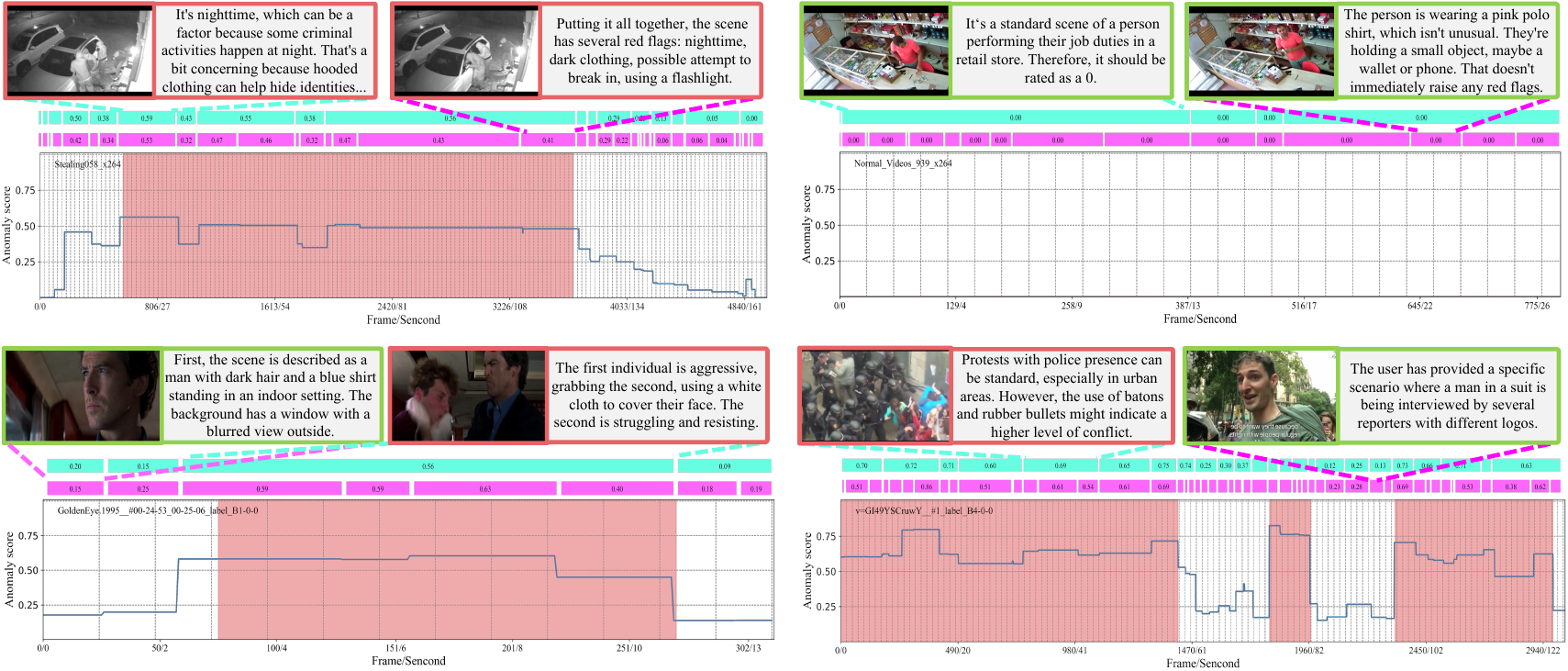} 
\caption{Qualitative results from \methodshort~on four test videos: two from UCF-Crime (top row) and two from XD-Violence (bottom row). 
The hierarchical video segment representations and corresponding anomaly scores are visualized alongside their key language explanations, with \inlineColorbox{Mycyan}{cyan} and \inlineColorbox{Myrose}{rose} rectangles denoting coarse cluster and fine cluster nodes respectively.
Each video's final anomaly scores (\setulcolor{blue!80}\ul{blue solid line}) are computed by inter-cluster node correlation. 
Ground-truth anomalies are highlighted by \inlineColorbox{red!35}{red} regions.} 
\label{fig:demo} 
\end{figure}

Figure~\ref{fig:demo} presents qualitative results of \methodshort~using sample videos from UCF-Crime and XD-Violence. 
Benefiting from its flexible granularity-aware tree video representation, \methodshort~accurately segments the boundaries between anomalous and normal events. Additionally, due to the scoring stability brought by inter-cluster node correlation, the anomaly scores generated by our method are overall smoother.
In addition, we observe that the same video can obtain divergent anomaly scores across its coarse and fine cluster representations. 
This divergence stems primarily from the lack of a uniform standard for the independent inference of nodes by VLMs and LLMs, which inevitably leads to scoring fluctuations.
Our inter-cluster correlation can reduce the impact of such instability.

\section{Conclusion}
\label{conclusion}
This paper presents a novel training-free framework \methodshort~for adaptive multi-granularity VAD. 
By constructing a hierarchical granularity-aware tree to achieve node-wise anomaly understanding and score refinement based on tree structure information,
our method overcomes the limitation of poor flexibility in anomaly detection based on fixed-length sliding temporal window sampling in existing methods.
The elimination of domain-specific training requirements and explicit explainability through pre-trained model reasoning make our framework particularly suitable for real-world surveillance applications.
\begin{ack}
This work was supported in part by the National Natural Science Foundation of China 62572387 and U22B2036, and Jiangsu Agricultural Science and Technology Innovation Fund (CX(24)3132), and Natural Science Basic Research Program of Shaanxi (Program No.2024JC-YBMS-498), and Shaanxi Provincial Key Research and Development Program - Key Project - Qinchuangyuan Original Innovation Window "Four Chains" Integration Project (2024PT-ZCK-93).
\end{ack}

\bibliographystyle{plain}
\bibliography{bib/vad}
\section*{NeurIPS Paper Checklist}

\begin{enumerate}

\item {\bf Claims}
    \item[] Question: Do the main claims made in the abstract and introduction accurately reflect the paper's contributions and scope?
    \item[] Answer: \answerYes{} 
    \item[] Justification: The abstract and introduction state the contributions of the paper.

\item {\bf Limitations}
    \item[] Question: Does the paper discuss the limitations of the work performed by the authors?
    \item[] Answer: \answerYes{} 
    \item[] Justification: Limitation Section is provided in the appendix.

\item {\bf Theory assumptions and proofs}
    \item[] Question: For each theoretical result, does the paper provide the full set of assumptions and a complete (and correct) proof?
    \item[] Answer: \answerYes{} 
    \item[] Justification: The proof is provided in the appendix.

    \item {\bf Experimental result reproducibility}
    \item[] Question: Does the paper fully disclose all the information needed to reproduce the main experimental results of the paper to the extent that it affects the main claims and/or conclusions of the paper (regardless of whether the code and data are provided or not)?
    \item[] Answer: \answerYes{} 
    \item[] Justification: We discuss the details required to reproduce the experiments in this paper in the Experiments and Appendix sections.

\item {\bf Open access to data and code}
    \item[] Question: Does the paper provide open access to the data and code, with sufficient instructions to faithfully reproduce the main experimental results, as described in supplemental material?
    \item[] Answer: \answerYes{} 
    \item[] Justification: We provide code in the appendix to reproduce our results, and link our code release in the anonymous github in section abstract.

\item {\bf Experimental setting/details}
    \item[] Question: Does the paper specify all the training and test details (e.g., data splits, hyperparameters, how they were chosen, type of optimizer, etc.) necessary to understand the results?
    \item[] Answer: \answerYes{} 
    \item[] Justification: We use standard datasets and splits, we provide hyperparameters in experimental details along with ablations in experiment sections and appendix to understand the contribution of each component in our algorithm.

\item {\bf Experiment statistical significance}
    \item[] Question: Does the paper report error bars suitably and correctly defined or other appropriate information about the statistical significance of the experiments?
    \item[] Answer: \answerYes{}{} 
    \item[] Justification:  Considering the limitation of computing resources, we repeated the main experiments and reported the mean and standard deviation in the appendix.

\item {\bf Experiments compute resources}
    \item[] Question: For each experiment, does the paper provide sufficient information on the computer resources (type of compute workers, memory, time of execution) needed to reproduce the experiments?
    \item[] Answer: \answerYes{} 
    \item[] Justification: We have provided details about compute resources used in supplemental material.
    
\item {\bf Code of ethics}
    \item[] Question: Does the research conducted in the paper conform, in every respect, with the NeurIPS Code of Ethics \url{https://neurips.cc/public/EthicsGuidelines}?
    \item[] Answer: \answerYes{} 
    \item[] Justification: We have read the ethics guidelines and confirm that we do not use human subjects, use existing datasets, explicitly discuss social impacts.

\item {\bf Broader impacts}
    \item[] Question: Does the paper discuss both potential positive societal impacts and negative societal impacts of the work performed?
    \item[] Answer: \answerYes{} 
    \item[] Justification: The potential social impact of this paper is discussed in the appendix.
    
\item {\bf Safeguards}
    \item[] Question: Does the paper describe safeguards that have been put in place for responsible release of data or models that have a high risk for misuse (e.g., pretrained language models, image generators, or scraped datasets)?
    \item[] Answer: \answerNA{} 
    \item[] Justification: No new datasets or generative models are released, and public datasets are used, so no protection measures are required.

\item {\bf Licenses for existing assets}
    \item[] Question: Are the creators or original owners of assets (e.g., code, data, models), used in the paper, properly credited and are the license and terms of use explicitly mentioned and properly respected?
    \item[] Answer: \answerYes{} 
    \item[] Justification:  UCF-Crime and XD-Violence are cited appropriately. 

\item {\bf New assets}
    \item[] Question: Are new assets introduced in the paper well documented and is the documentation provided alongside the assets?
    \item[] Answer: \answerYes{}, 
    \item[] Justification: We provide code and instructions.

\item {\bf Crowdsourcing and research with human subjects}
    \item[] Question: For crowdsourcing experiments and research with human subjects, does the paper include the full text of instructions given to participants and screenshots, if applicable, as well as details about compensation (if any)? 
    \item[] Answer: \answerNA{} 
    \item[] Justification: This paper does not involve crowdsourcing nor research with human subjects.

\item {\bf Institutional review board (IRB) approvals or equivalent for research with human subjects}
    \item[] Question: Does the paper describe potential risks incurred by study participants, whether such risks were disclosed to the subjects, and whether Institutional Review Board (IRB) approvals (or an equivalent approval/review based on the requirements of your country or institution) were obtained?
    \item[] Answer: \answerNA{} 
    \item[] Justification: This article does not involve subjects and related approval requirements.

\item {\bf Declaration of LLM usage}
    \item[] Question: Does the paper describe the usage of LLMs if it is an important, original, or non-standard component of the core methods in this research? Note that if the LLM is used only for writing, editing, or formatting purposes and does not impact the core methodology, scientific rigorousness, or originality of the research, declaration is not required.
    \item[] Answer: \answerYes{} 
    \item[] Justification: Our core methodology integrates large language models (LLMs) as a non-standard component for semantic anomaly reasoning in video analysis.
\end{enumerate}

\addcontentsline{toc}{section}{Appendix} 
\renewcommand \thepart{} 
\renewcommand \partname{}

\part{
    \centering{\Large{\textbf{\articletitle}}}\\
}
\centerline{\Large{{Technical Appendices}}}

\parttoc

\newpage

\appendix



The appendix begins by detailing the algorithmic process underlying the construction of the HGTree and establishing the proof of its representational completeness. Furthermore, it presents additional experimental details, including extensive ablation studies and comparative analyses. Finally, the appendix examines the limitations of this work and its potential societal impact.
\section{Hierarchical Granularity-aware Tree}

\subsection{TreeInit: Granularity-Aware Binary Tree Construction}\label{app:hgtree}

\begin{algorithm}[h!]
\caption{TreeInit: Granularity-Aware Binary Tree Construction Algorithm (Section~\ref{sec:3.1.2})}
\label{alg:tree_init}
\begin{algorithmic}[1]
\Require 
    \Statex Video $V_{1:T}$ with $T$ frames, 
    \Statex Confidence scores $\hat{C} = \{(\tau_i, \hat{c}_i)\}_{i=1}^N$,
    \Statex Confidence threshold $\gamma_{\text{min}}$
\Ensure Binary tree $\mathcal{T} = \{([s_j, e_j], [\hat{c}_s^j, \hat{c}_e^j])\}_{j=1}^M$

\State $\mathcal{T} \gets \emptyset$, $\mathcal{U} \gets \emptyset$ 
    \Comment{Result set \& consumed split points}
\State Push root node $\mathcal{D} \gets \big[[1, T]\big]$ 
    \Comment{DFS stack initialization}

\While{$\mathcal{D} \neq \emptyset$}
    \State $[l, r] \gets \mathcal{D}.\texttt{pop}()$
    
    \State $\hat{c}_l \gets \mathbb{I}(l=1) \cdot 1 + \mathbb{I}(l>1) \cdot \hat{c}_l$ \Comment{Left boundary confidence}
    \State $\hat{c}_r \gets \mathbb{I}(r=T) \cdot 1 + \mathbb{I}(r<T) \cdot \hat{c}_r$ \Comment{Right boundary confidence}

    \State $\mathcal{T}.\texttt{add}\left([l, r], [\hat{c}_l, \hat{c}_r]\right)$
    
    \State \textbf{Find split} $\tau^* \gets \underset{\tau \in (\Psi \setminus \mathcal{U}) \cap (l, r)}{\arg\max}\ \hat{c}_\tau$
    \Comment{Select the highest remaining confidence point}
    
    \If{$ \hat{c}_{\tau^*} \geq \gamma_{\text{min}}$}
        \State $\mathcal{U}.\texttt{add}(\tau^*)$
        \State $\mathcal{D}.\texttt{push}([\tau^*, r])$ 
            \Comment{Right child}
        \State $\mathcal{D}.\texttt{push}([l, \tau^*])$ 
            \Comment{Left child}
    \EndIf
\EndWhile

\State \Return $\operatorname{Sort}(\mathcal{T}, l_j \uparrow)$ 
    \Comment{Sort by start time}
\end{algorithmic}
\end{algorithm}

\subsection{Proof of Coverage Completeness in Hierarchical Coarse-Fine Clustering}\label{app:proof}
\begin{theorem}[Coverage Completeness] \label{thm:coverage}
Based on the method described in Section~\ref{sec:3.1.3}, we get $\mathcal{T}' = (\mathcal{S}_{coarse}', \mathcal{S}_{fine}')$, where $|\mathcal{S}_{coarse}'| = M_c'$~ and ~$|\mathcal{S}_{fine}'| = M_f' $ . Then:

\textbf{
The original video sequence $V_{1:T}$ can be exactly reconstructed through temporal concatenation of segments from either the coarse cluster $\mathcal{S}_{\text{coarse}}'$ or the fine cluster $\mathcal{S}_{\text{fine}}'$:}
\begin{equation}\label{eq:completeness}
\bigcup_{\mathcal{N}_i \in \mathcal{S}_{coarse}'}[l_i, r_i] = [1, T], \bigcup_{\mathcal{N}_i \in \mathcal{S}_{fine}'}[l_i, r_i] = [1, T].
\end{equation}
\end{theorem}

\emph{Notations}:
\begin{itemize}
    \item $\mathcal{N}_i = ([l_i, r_i], [\hat{c}_l^{(i)}, \hat{c}_r^{(i)}])$: 
    A tree node represents a generalized event video segment, with boundary frames and their confidences as $[l_i, ri]$ and $[\hat{c}_l^{(i)}, \hat{c}_r^{(i)}]$ respectively.
    \item $\prec$: Parent-child relation in $\mathcal{T}$ ($\mathcal{N}_j \prec \mathcal{N}_i \iff \mathcal{N}_i$ is a child of $\mathcal{N}_j$)
    \item $\mathcal{T}_{\textit{leaf}} \triangleq \{\mathcal{N}_i \in \mathcal{T} \,|\, \nexists~ \mathcal{N}_j \prec \mathcal{N}_i\}$: Leaf node set of $\mathcal{T}$
\end{itemize}

\begin{proof}

\noindent\textbf{Part 1: Initial Coverage Guarantee}
The root node $\mathcal{N}_0 = ([1, T], [1, 1]) \in \mathcal{T}$ spans the full video by definition. Through iterative splitting in Algorithm~\ref{alg:tree_init}, each parent node $\mathcal{N}_p = ([l_p, r_p], [\hat{c}_l^{(p)}, \hat{c}_r^{(p)}])$ is partitioned into non-overlapping child nodes:
\begin{equation} \label{eq:split}
\mathcal{N}_c^L = ([l_p, \tau^*], [\hat{c}_l^{(c)}, \hat{c}_{\tau^*}^{(c)}]), \quad \mathcal{N}_c^R = ([\tau^*, r_p], [\hat{c}_{\tau^*}^{(c)}, \hat{c}_r^{(c)}])
\end{equation}
where $\tau^* \in (l_p, r_p)$. This implies: 
\begin{equation}\label{eq:leaf}
\bigcup_{\mathcal{N}_i \in \mathcal{T}_{leaf}} [l_i, r_i] = [1, T]
\end{equation}

\noindent\textbf{Part 2: Coarse Cluster Guarantee}
The $\operatorname{RemoveDup}$ operator filters nodes through:
\begin{equation} \label{eq:removedup}
\mathcal{S}_{\textit{coarse}}' = \{\mathcal{N}_i \in \mathcal{S}_{\textit{coarse}} \,|\, \nexists \ \mathcal{N}_i \prec \mathcal{N}_j \}
\end{equation}
As the above operation exclusively targets non-leaf nodes in $\mathcal{S}_{coarse}$ and leaves leaf nodes unchanged.
Therefore, the current leaf nodes satisfies the expression completeness of the original video shown in Eq.~\ref{eq:leaf}, and then satisfies the first item of Eq.~\ref{eq:completeness}: $\bigcup_{\mathcal{N}_i \in \mathcal{S}_{coarse}'}[l_i, r_i] = [1, T]$.

\noindent\textbf{Part 3: Fine Cluster Guarantee} 

The $\operatorname{Complete}$ operator ensures coverage via two mechanisms:

1. Boundary alignment: For edge cases:
\begin{align}\label{eq:boundary}
\text{if } \min_{\mathcal{N}_i \in \mathcal{S}_{\textit{fine}}'} l_i > 1 &: \text{insert } \mathcal{N}_1 \text{ from } \mathcal{S}_{\textit{coarse}}'  \\
\text{if } \max_{\mathcal{N}_i \in \mathcal{S}_{\textit{fine}}'} r_i < T &: \text{append } \mathcal{N}_{ M_c'} \text{ from } \mathcal{S}_{\textit{coarse}}' \nonumber
\end{align}

2. Bridge the gap between nodes: For any adjacent nodes $\mathcal{N}_i = ([l_i, r_i], [\hat{c}_l^{(i)}, \hat{c}_r^{(i)}])$ and $\mathcal{N}_{i+1} = ([l_{i+1}, r_{i+1}], [\hat{c}_l^{(i+1)}, \hat{c}_r^{(i+1)}])$ in $\mathcal{S}_{fine}'$ with $r_i < l_{i+1}$:
\begin{equation} \label{eq:gap}
\exists ~\{\mathcal{N}_c\} \subset \mathcal{S}_{coarse}' \text{ s.t. } \bigcup_{c}[l_c, r_c] = [r_i, l_{i+1}]
\end{equation}

Through Eq.~\ref{eq:boundary} and Eq.~\ref{eq:gap}, the second term of Eq.~\ref{eq:completeness} is satisfied: $\bigcup_{\mathcal{N}_i \in \mathcal{S}_{fine}'}[l_i, r_i] = [1, T]$.



\noindent\textbf{Conclusion}: Both coarse and fine cluster maintain complete temporal coverage through Section~\ref{sec:3.1.3} process.
\end{proof}

\section{Generic Event-centric Anomaly Scoring and Refining}
\subsection{Prior-infused Node Scoring}


This section mainly supplements the prompt details used by VLM and LLM.
First, by employing $P_b \circ P_c$ (as demonstrated in Section~\ref{sec:prior-gen}), we input the prompt into the LLM~\cite{deepseekai2025deepseekr1}~\footnote{\url{https://chat.deepseek.com/}} to derive prior knowledge that excludes ill-posed semantic cues.
The prior knowledge $B$ is shown in Table~\ref{tab:prior-ucf} and Table~\ref{tab:prior-xd} respectively.

The model configuration details of the VLM for describing video content and the LLM for scoring anomalies are consistent with their open-source repositories~\footnote{\url{https://huggingface.co/lmms-lab/LLaVA-Video-7B-Qwen2}}~\footnote{\url{https://huggingface.co/deepseek-ai/DeepSeek-R1-Distill-Qwen-14B}}.

\subsubsection{Multidimensional Prior Knowledge Generation Prompt}\label{sec:prior-gen}

\paragraph{UCF-Crime}
\textit{
"To help video anomaly detection agent review the occurrence of abnormal events, it is now necessary to pre-analyze
possible anomalies to establish a prior knowledge base that matches abnormal events. The video taken has no sound, and may have a long distance or a blurry picture. There may be Abuse, Arrest, Arson, Assault, Burglary, Explosion, Fighting, RoadAccidents, Robbery, Shooting, Shoplifting, Stealing and Vandalism 13 types of events. Please carefully analyze these scenes. Then point out the characteristics of each abnormal event from the following three perspectives: the scene environment, characters or specific objects, actions or behaviors that occurred."}

\paragraph{XD-Violence}
\textit{
“To help video anomaly detection agent review the occurrence of abnormal events, it is now necessary to pre-analyze possible anomalies to establish a prior knowledge base that matches abnormal events. The video taken has no sound, and may have a long distance or a blurry picture. There may be Abuse,  Explosion, Fighting, Car Accident, Shooting and Riot 6 types of events. Please carefully analyze these scenes. Then point out the characteristics of each abnormal event from the following three perspectives: the scene environment, characters or specific objects, actions or behaviors that occurred.”
}

\subsubsection{Multidimensional Prior Knowledge}
The prior knowledge $B$ generated for the UCF-Crime and XD-Violence datasets are shown in Table~\ref{tab:prior-ucf} and Table~\ref{tab:prior-xd} respectively.
\paragraph{UCF-Crime} 
\begin{xltabular}{\textwidth}{c X X X}
\caption{Multidimensional Prior Knowledge of UCF-Crime Dataset.}
\label{tab:prior-ucf}\\
\toprule
\textbf{Abnormal Event Type} & \textbf{Scene Environment Features} & \textbf{Character/Object Features} & \textbf{Action/Behavior Features} \\
\midrule
\endfirsthead
    Abuse & 
    Secluded spaces (indoors/corners), non-public areas (private locations) & 
    Two parties in physical conflict (perpetrator/victim), dragging tools (ropes/clubs) & 
    Shoving/dragging, repeated hitting, restraining movement (pinning down) \\ 
    \midrule
    Arrest & 
    Public areas (streets/squares), zones with police vehicles or officers & 
    Uniformed police, handcuffs, batons or firearms & 
    Forced restraint, frisking, escorting to vehicles, lying on the ground \\ 
    \midrule
    Arson & 
    Areas with flammable materials (warehouses/vehicles), abnormal smoke/flames & 
    Individuals holding flammable containers (gasoline bottles), ignition tools (lighters) & 
    Throwing incendiary objects, fleeing quickly, repeatedly checking the fire \\ 
    \midrule
    Assault & 
    Narrow passages, crowded areas with sudden dispersion (subway stations/bar entrances) & 
    Armed individuals (knives/blunt weapons), victims struggling on the ground & 
    Sudden lunging, weapon swinging, victims adopting defensive postures \\ 
    \midrule
    Burglary & 
    Damaged doors/windows, unlit buildings at night, surveillance blind spots (back alleys) & 
    Masked/dark-clothed individuals, lock-picking tools (pliers), backpacks (for loot) & 
    Peering through windows, picking locks, rummaging through items \\ 
    \midrule
    Explosion & 
    Smoke spreading, flying debris, crowds fleeing outward from a central point & 
    Suspicious packages/vehicles, post-explosion wreckage (metal fragments) & 
    Throwing motions, sudden flash of flames, crowds crouching/running \\ 
    \midrule
    Fighting & 
    Public spaces (restaurants/stadiums) with concentrated physical conflicts, overturned furniture & 
    Multiple people entangled, bleeding faces, torn clothing & 
    Punching/kicking, hair-pulling, siege \\ 
    \midrule
    Road Accidents & 
    Collision points (intersections/curves), skid marks, scattered debris, traffic congestion & 
    Deformed vehicles, deployed airbags, paramedics (uniforms/stretchers) & 
    Sudden braking, vehicle rollovers, pedestrians being hit \\ 
    \midrule
    Robbery & 
    Streets/ATM areas, fast-moving vehicles (motorcycles/cars) & 
    Threats with guns/knives, motorcycle helmets (face concealment), stolen items (bags) & 
    Snatching and fleeing, threatening gestures, vehicles abruptly stopping/accelerating \\ 
    \midrule
    Shooting & 
    Crowds suddenly ducking/fleeing, vehicles braking abruptly, bullet holes in windows & 
    Gun-wielding individuals, gunshot victims falling, spent shell casings & 
    Aiming firearms, continuous firing, seeking cover \\ 
    \midrule
    Shoplifting & 
    Loitering near shelves, surveillance blind spots (corners), suspicious concealment (coats) & 
    Frequently observing staff, hiding items (in bags/under clothing) & 
    Concealing items in clothing, glancing around nervously, quickly leaving shelves \\ 
    \midrule
    Stealing & 
    Crowded areas (subways/markets), sudden disappearance of target items (wallets/phones) & 
    Close proximity to victims, distractions (e.g., bumping), rapid transfer of stolen goods & 
    Pickpocketing (hands reaching into pockets), passing loot to accomplices \\ 
    \midrule
    Vandalism & 
    Graffiti-covered walls, shattered glass, toppled public facilities (trash cans/fences) & 
    Spray paint cans, hammers/stones, targets (cameras/glass) & 
    Smashing motions, spraying walls, kicking facilities \\ 
    \midrule
    \multicolumn{4}{@{}p{\textwidth}@{}}{\footnotesize
    \textbf{Table Notes:}
    \begin{enumerate}[leftmargin=*,nosep]
    \item Scene environment features capture spatial anomalies (e.g., secluded corners) and physical damage patterns
    \item Character/object features focus on suspicious entities and high-risk items
    \item Action/behavior features characterize motion dynamics critical for low-quality video analysis
    \end{enumerate}
    \textbf{Recognition Tips:}
    \begin{enumerate}[leftmargin=*,nosep]
    \item Blurry footage: Track group behavior changes (crowd fleeing patterns)
    \item Long-distance: Monitor environmental dynamics (smoke/glass shattering)
    \item Silent videos: Analyze action intensity (repeated hitting motions)
    \end{enumerate}
    }\\
    \bottomrule
\end{xltabular}

\paragraph{XD-Violence} 

\begin{xltabular}{\textwidth}{c X X X}
\caption{Multidimensional Prior Knowledge of XD-Violence Dataset.}
\label{tab:prior-xd}\\
\toprule
\textbf{Abnormal Event Type} & \textbf{Scene Environment Features} & \textbf{Character/Object Features} & \textbf{Action/Behavior Features} \\
\midrule
\endfirsthead

Abuse & 
Secluded or private settings (alleyways, dimly lit rooms), lack of bystanders & 
Dominant/submissive individuals with indistinct blunt objects (belts, sticks) & 
Sudden aggressive movements (hitting/grabbing), victim recoiling/fleeing, prolonged physical contact \\ 
\midrule

Explosion & 
Sudden bright flash with smoke/fire, structural damage (collapsed walls) & 
Chaotically moving people, objects near blast source (vehicles, trash bins) & 
Rapid light/smoke expansion, crowd scattering, lingering smoke/flames \\ 
\midrule

Fighting & 
Public areas with bystanders (streets, bars), overturned furniture & 
Aggressive postures, broken objects nearby (bottles, chairs) & 
Repetitive punching/kicking, high-intensity movements, bystander reactions \\ 
\midrule

Car Accident & 
Roads with sudden traffic stops, skid marks/debris (broken glass) & 
Collided vehicles (distorted shapes), injured individuals & 
Rapid deceleration/impact, post-crash behaviors (crowd gathering, emergency services) \\ 
\midrule

Shooting & 
Public spaces with visible escape routes (malls, streets) & 
Shooter's stance (arm extended), collapsing victims & 
Crowd panic (ducking/sprinting), post-shooting patterns (bodies, law enforcement) \\ 
\midrule

Riot & 
Large crowds with fires/smoke, broken infrastructure (smashed windows) & 
Protesters with makeshift weapons, riot police/armored vehicles & 
Group violence (throwing objects), chaotic dispersal patterns (wave-like movements) \\ 
\midrule

\multicolumn{4}{@{}p{\textwidth}@{}}{\small
\textbf{Low-quality Analysis Guidelines:}
\begin{itemize}
\item Focus on movement patterns (sudden scattering/aggressive postures)
\item Detect environmental changes (smoke/debris) over fine details
\item Analyze group dynamics (crowd panic/clustering)
\item Prioritize spatiotemporal anomalies (light flashes/vehicle stops)
\end{itemize}
} \\
\bottomrule
\end{xltabular}

\subsubsection{Video Content Description Prompt}
If directly driving the VLM to describe the video, its prompt $P_d$ is as follows:

\textit{
"Please describe the content in the video in detail."}

If prior knowledge $B$ is incorporated, the form of $B \circ P_d$ is adopted.

\subsubsection{Abnormal Scoring Prompt}
The prompt $P_s$ used in this section is divided into two parts: Scoring Criteria and Output Format Control.
\paragraph{Scoring Criteria}
\textit{"How would you rate the scene described on a scale from 0 to 1, with 0 representing a standard scene and 1 denoting a scene with suspicious or potentially criminal activities?"
}
\paragraph{Output Format Control}
\textit{"Please provide the response in the form of a Python list and respond with only one number in the provided list below [0, 0.1, 0.2, 0.3, 0.4, 0.5, 0.6, 0.7, 0.8, 0.9, 1.0] without any textual explanation. It should begin with `[' and end with `]'."
}

\section{More Results}\label{sec:app_results}
\subsection{More Experimental Details}
Based on the experimental details described in Section~\ref{sec:details}, the $\gamma_{min}=0.4$ and K-Means clustering algorithm are used to generate the HGTree for inference.
In the inter-cluster node refinement process, 
we implemented a top-K control for the final weighted neighborhood node numbers. Additionally, this process also includes the temperature parameter $\tau$ of softmax.
In the Inter-cluster Node Correlation, 
the hyperparameter $\beta$ affects the weight of coarse and fine clusters in the final anomaly score.


\subsection{Effect of Different GEBD Methods}
As shown in Section~\ref{sec:3.1}, we suppress the negative impact of low-quality generalized event boundaries on VAD performance in several ways.
In addition, we select different GEBD models trained on Kinetics-GEBD dataset [59] to demonstrate the stability of the above strategies. Table~\ref{table:app_GEBD} reveals a direct correlation between their VAD performance on UCF-Crime and their original Kinetics-GEBD~\cite{shou2021generic} dataset results. This confirmation highlights two key findings: (1) The quality of GEBD models remains an influential factor, as improved GEBD implementations consistently yield better performance; (2) Our architectural innovations demonstrate robust adaptability to boundary quality variations.

From a domain shift perspective, Kinetics-GEBD's open-world diversity provides transferable representations superior to those of constrained anomaly datasets (UCF-Crime, XD-Violence, and MSAD). This GEBD-based pre-training aligns with established transfer learning paradigms, boosting cross-domain detection robustness.

\begin{table}[h]
\centering
\caption{Results of VADTree based on different GEBD models on UCF-Crime dataset.}
\label{table:app_GEBD}
\resizebox{\textwidth}{!}{%
\begin{tabular}{l ccc}
\toprule
\textbf{GEBD Method} & \textbf{Kinetics-GEBD-Val F1(\%)} & \textbf{Kinetics-GEBD-Test F1(\%)} & \textbf{UCF-Crime AUC(\%)} \\
\midrule
SceneDetect\tablefootnote{\url{https://github.com/Breakthrough/PySceneDetect}} & - & - & 80.00 \\
BasicGEBD-ResNet50 & 73.70 & 76.80 & 82.85 \\
EfficientGEBD-ResNet18-L4 & 78.20 & - & 84.70 \\
EfficientGEBD-ResNet50-L4 (Ours) & \textbf{78.64} & \textbf{78.70} & \textbf{84.74} \\
\bottomrule
\end{tabular}%
}
\end{table}

\subsection{Effect of Intra-cluster Node Refinement Configuration}
The neighborhood size parameter K in Eq.~\ref{eq:refine} governs the trade-off between localized feature precision and noise suppression. 
Table~\ref{table:app_K} demonstrates substantial AUC gains from neighborhood node refinement: 9.88\% for VADTree-Fine and 7.14\% for VADTree-Coarse when expanding K from 0 to 10. Both clusters exhibit maximal improvements within this critical initialization range.
Performance stabilizes between K=10 and K=15, with the fine and coarse clusters maintaining AUC of 83.03–83.05\% and 82.55–82.81\%, respectively. A gradual degradation when K exceeds 15 indicates that the optimal balance between contextual integration and noise suppression has been reached within this range.

\begin{table}[h]
    \caption{Influence of top-\textbf{K} weighted neighborhood nodes on AUC (\%).}
    \label{table:app_K}
	\centering
	\footnotesize
	\begin{tabular}{c| ccc ccc }
        \toprule
		\textbf{K} & \textbf{0} &\textbf{5} & \textbf{10}& \textbf{15} & \textbf{20} & \textbf{25} \\
         \midrule
		VADTree-Coarse& 75.67 & 81.84& 82.81& 82.55& 81.96& 81.73\\
		VADTree-Fine & 73.17 & 79.77& 83.05& 83.03& 82.65& 82.43\\
        \bottomrule
	\end{tabular}
\end{table}
The temperature coefficient $\tau$ in Eq.~\ref{eq:refine} regulates the entropy characteristics of Softmax-derived distributions while maintaining ordinal relationships between elements. 
Our empirical analysis (Table~\ref{table:app_tau}) reveals that as $\tau$ approaches zero ($\tau=0.001$), the distribution collapses into a degenerate form concentrated solely on the maximal element, equivalent to a non-weighted selection. 
Progressively increasing the parameter to moderate values produces an AUC plateau of 83.05\% for the VADTree-Fine exhibiting minimal variance.
Notably, excessive temperature values ($\tau=100$) induce uniform distributions, degrading performance to 82.43\% AUC for VADTree-Fine. 
This analysis indicates that the optimal parameter range is $\tau \in [0.01,1]$, where an optimal balance is achieved between distribution sharpness and model stability.
For the experiments of VADTree-Coarse, we can get similar conclusions.
\begin{table}[h]
\caption{Influence of softmax temperature $\tau$ on AUC (\%).}
\label{table:app_tau}
	\centering
	\footnotesize
	\begin{tabular}{c| ccc ccc }
        \toprule
		$\bm{\tau}$ & \textbf{0.001} & \textbf{0.01}& \textbf{0.1} & \textbf{1} & \textbf{10} &\textbf{100}  \\
         \midrule
		VADTree-Coarse& 78.72 & 80.68& 82.81& 82.42& 82.21& 82.20\\
		VADTree-Fine& 77.83 & 80.72& 83.05& 83.05& 83.02& 83.02\\
        \bottomrule
	\end{tabular}

\end{table}
\subsection{Effect of Inter-cluster Node Correlation Configuration}
\begin{figure}[h!]
  \centering
  \begin{minipage}[t]{0.43\textwidth} 
    \centering
    \makeatletter\def\@captype{table}\makeatother
    \caption{Performance comparison demonstrating the efficacy of inter-cluster correlation. The integration of hierarchical clusters in VADTree yields the highest AUC.}
    \label{tab:app_correlation }
    \resizebox{\textwidth}{!}
    {
    \begin{tabular}{l ccc}
        \toprule
        \textbf{Datasets} & \textbf{UCF-Crime} & \textbf{XD-Violence} & \textbf{MSAD} \\
        \midrule
        VADTree-Fine & 83.05 & 90.04 & 86.71 \\
        VADTree-Coarse & 82.81 & 89.36 & 87.01 \\
        VADTree & \textbf{84.74} & \textbf{90.44} & \textbf{89.32}\\
        \bottomrule
    \end{tabular}
    }
  \end{minipage}
  \hfill
  \begin{minipage}[t]{0.55\textwidth} 
      \centering
    \caption{Influence of inter-cluster-correlation control coefficient $\beta$ on AUC.}\label{fig:app_beta}
      \includegraphics[
      width=\textwidth]{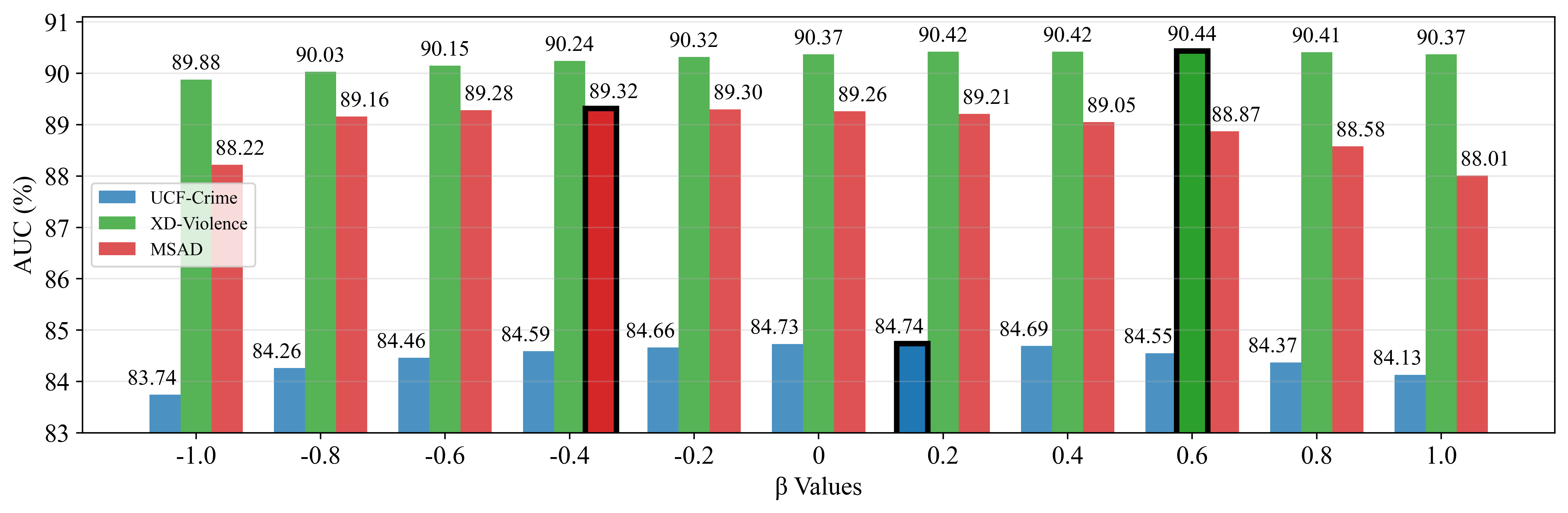}
  \end{minipage}
\end{figure}

The $\beta$ coefficient regulates parent-child node interplay in our cohesion-driven correlation (Eq.~\ref{eq:final score}).  Notably, when $\beta = 0$, the correlation operation degenerates to a simple average of anomaly scores from parent and child nodes.
As quantified in Figure~\ref{fig:app_beta}, the optimal control coefficient~$\beta=0.4$ delivers peak AUC performance at 84.74\% for UCF-Crime dataset, indicating an effective equilibrium between parent node contextual integration and child nodes semantic specificity.
Additionally, limited AUC fluctuation demonstrates the hierarchy's inherent noise suppression capability.  
This validates our weighted design as an effective strategy for multi-granularity fusion.

\subsection{More Ablation Experiments}
Our additional ablation analysis examines the contribution of each component in \methodshort~, namely the HGTree fine cluster, prior-infused node scoring, intra-cluster node refinement, and inter-cluster node correlation, to assess their individual impact on performance.
We also evaluate the effectiveness of our components on the 10s fixed-length sliding temporal window (TW) sampling method. Table~\ref{tab:app_ablation_components} shows the results of all ablated variants of \methodshort. The experiment shows that each component has a significant impact on our final results. At the same time, these components are still effective for methods using fixed-length sliding temporal window sampling.
\begin{table*}[t]
    \centering 
    \caption{Ablation study of \methodshort~components on the UCF-Crime dataset. The upper and lower panels present experiments using HGTree and 10 seconds fixed-length sliding temporal window (TW) sampling respectively.}
    \label{tab:app_ablation_components}
    \footnotesize
        {%
        
            \begin{tabular}{cccc|c}
            \toprule
            \textbf{HGTree Fine Cluster} & \textbf{Prior-infused $f_{\text{VLM}}$} & \textbf{Refinement}& \textbf{Correlation}& \textbf{AUC (\%)} \\
            \midrule
                \textcolor{green}{\ding{51}} &     \textcolor{red}{\ding{55}}   & \textcolor{green}{\ding{51}}  & \textcolor{green}{\ding{51}} & 83.08\\
                \textcolor{green}{\ding{51}} &  \textcolor{green}{\ding{51}} &   \textcolor{red}{\ding{55}}  & \textcolor{green}{\ding{51}}  &77.97 \\
                \textcolor{green}{\ding{51}} & \textcolor{green}{\ding{51}} & \textcolor{green}{\ding{51}} &\textcolor{red}{\ding{55}}  & 83.05 \\
                \textcolor{green}{\ding{51}} & \textcolor{green}{\ding{51}} & \textcolor{green}{\ding{51}} &\textcolor{green}{\ding{51}} & \textbf{84.74} \\
            \midrule
                \textcolor{red}{\ding{55}} &     \textcolor{red}{\ding{55}}   & \textcolor{red}{\ding{55}}  & \textcolor{red}{\ding{55}} & 72.93\\
                \textcolor{red}{\ding{55}}      & \textcolor{green}{\ding{51}} & \textcolor{red}{\ding{55}} & \textcolor{red}{\ding{55}}& 75.21 \\
                \textcolor{red}{\ding{55}}      & \textcolor{red}{\ding{55}} & \textcolor{green}{\ding{51}} & \textcolor{red}{\ding{55}}&80.62  \\
                \textcolor{red}{\ding{55}}      & \textcolor{green}{\ding{51}} & \textcolor{green}{\ding{51}} & \textcolor{red}{\ding{55}}& 82.81 \\
            \bottomrule
            \end{tabular}%
        }
            
\end{table*}
\subsection{Comparison of \methodshort~and Different Video Sampling Methods}
In this experiment, we conducted a comparative analysis of \methodshort~against mainstream video sampling approaches, focusing on final anomaly detection performance and computational efficiency. The fixed-length sliding temporal window (TW) method, employed by LAVAD and VERA~\cite{zanella2024harnessing,ye2024vera}, serves as our primary comparison method. Additionally, we propose three metrics to evaluate video sampling efficiency: (1) Number of Segments (NoS), defined as the total number of video segments sampled from the test dataset; (2) Mean Intersection over Union (mIoU), computed by first identifying the maximum temporal IoU between each anomalous event and all sampled segments within a video, then averaging these maximum values across all events;
(3) Mean Intersection Frames (mIF), similar to mIoU, replaces the IoU metric with the number of intersecting frames between sampled video segments and ground truth abnormal segments.

\begin{table*}[h!]
    \caption{Results of 
    \methodshort~variants with different video sampling methods on the UCF-Crime Dataset. 16f represents a stride of 16 frames. \textbf{NoS} indicates the number of generated video segments. \textbf{mIoU} and \textbf{mIF} are used to measure the quality of video sampling.
    }
    \label{tab:sampler}
    \centering
    \footnotesize
        {%
            \begin{tabular}{ccc|ccc|c}
            \toprule
            \textbf{Method} & \textbf{TW Length} & \textbf{Stride}& \textbf{NoS}$\downarrow$&\textbf{mIoU}$\uparrow$&\textbf{mIF}$\uparrow$& \textsc{AUC (\%)} \\

            \midrule
                sliding TW & 5s & 5s  &7558 &0.41 &122& 82.06\\
                sliding TW & 10s & 10s  &3852 &0.40&191 &82.81 \\
                sliding TW & 20s & 20s  &1994  &0.33&265   &81.33  \\
                sliding TW & 10s & 16f~\cite{zanella2024harnessing,ye2024vera} &69634 &0.51& 210& 82.87
                \\
                \methodshort-Coarse & - &- &2248 & 0.37&369 & 82.81         \\%
                \methodshort-Fine & - &- &6365 & 0.40&233   & 83.05         \\%
                \methodshort & - &- &8613 & 0.47&343        & \textbf{84.74} \\%
                \methodshort~+ Redundant & - &- &12440 & 0.52&456        & - \\%
            \bottomrule
            \end{tabular}%
        }
\end{table*}

As demonstrated in Table~\ref{tab:sampler}, non-overlapping implementations of the TW strategy exhibit poor alignment with anomalous events. 
While dense overlapping sampling with short strides (16 frames)~\cite{zanella2024harnessing,ye2024vera} marginally improves AUC ROC it produces 8× more segments than~\methodshort, incurring significant computational costs without commensurate performance benefits. 
Our proposed \methodshort~achieves superior anomaly detection performance while maintaining comparable computational efficiency to non-overlapping TW baselines, demonstrating effective balance between precision and resource utilization.
\subsection{Computational Analysis}
According to the performance report by LAVAD, its VLM Caption module integrates the results of five BLIP-2 models. The parameter counts used by LAVAD are 3.6 times that of our method VADTree (as shown in Table~\ref{tab:parameter}) .
We display the total inference time (GPU hours) of LAVAD and VADTree on two NVIDIA GeForce RTX 3090 GPUs in Table~\ref{tab:consumption}. The time consumption of the VLM Caption, LLM Summary, and LLM Scoring parts of LAVAD is estimated.

A closer examination of Table~\ref{tab:consumption} reveals the following key observations: (1) VADTree-Coarse requires less than 30\% of LAVAD's GPU hours (16.5 vs. 55.9) while achieving a 2.53\% higher AUC (82.81 vs. 80.28) on UCF-Crime. This confirms that our method achieves a better trade-off between computational efficiency and detection accuracy compared to LAVAD. (2) Our VADTree framework is highly flexible, with the core HGTree construction process being computationally efficient. Both the VLM and LLM components are modular, allowing for adjustments based on computational constraints. Importantly, high-cost inference models are not essential for VADTree's effectiveness. (3) The default VADTree’s inference time is primarily influenced by the reasoning phase of DeepSeek-R1-Distill-Qwen-14B-think in the LLM scoring module. Replacing it with faster variants (e.g., DeepSeek-R1-Distill-Qwen-14B-no-think or t5gemma-9B-2B) significantly reduces inference time. We intentionally preserve the "Think" process because it generates valuable intermediate reasoning steps that significantly enhance anomaly interpretation. Even without this phase, our variants outperform LAVAD in AUC performance while maintaining lower inference times in all cases.
\begin{table}[h]
  \centering
    \caption{Component-level parameter analysis of VADTree and LAVAD.}
    \label{tab:parameter}
    \resizebox{0.99\textwidth}{!}
            {%
        \begin{tabular}{l|ccc ccc}
        \toprule 
        \textbf{Methods} & \textbf{\makecell{HGTree\\Construction}} & \textbf{\makecell{Video/Text\\Encoding}}  &  \textbf{\makecell{VLM\\Caption}} &  \textbf{\makecell{LLM\\Summary}} &  \textbf{\makecell{LLM\\Scoring} }& \textbf{Total}  \\
        \midrule
        LAVAD & - & ImageBind\textunderscore Huge-1.2B & 
        \makecell{OPT-6.7B $\times$ 2 + \\FLAN-T5XL-3B $\times$ 2 +\\FLAN-T5XXL-33B}
        & Llama-2-13B-chat & Llama-2-13B-chat & 79.6B \\
         VADTree & ResNet50-25.6M & ImageBind\textunderscore Huge-1.2B & LLaVA-NeXT-Video-7B & - & DeepSeek-R1-Distill-Qwen-14B & 22.2B \\
        \bottomrule
        \end{tabular}
            }
\end{table}%
    
\begin{table}[h]
  \centering
    \caption{Component-level inference time consumption analysis of VADTree and LAVAD on UCF-Crime dataset. The bold font indicates that VADTree's total GPU hours is lower than that of LAVAD. }
    \label{tab:consumption}
    \resizebox{0.99\textwidth}{!}
            {%
        \begin{tabular}{l|ccc ccc c}
        \toprule \textbf{Methods} & \textbf{\makecell{HGTree\\Construction}} & \textbf{\makecell{Video/Text\\Encoding}}  &  \textbf{\makecell{VLM\\Caption}} &  \textbf{\makecell{LLM\\Summary}} &  \textbf{\makecell{LLM\\Scoring} }& \textbf{\makecell{Total\\(GPU hours)}} & \textbf{AUC(\%)} \\
        \midrule
         LAVAD & - & 5.1h & 20h &7.7h $\times$ 2&7.7h $\times$ 2& 55.9 & 80.28 \\
         VADTree-Coarse & 0.3h & 0.2h &5.2h $\times$ 2& - &2.8h $\times$ 2& \textbf{16.5} & 82.81 \\
         VADTree-Fine & 0.3h & 0.4h &14.8h $\times$ 2& - &7.9h $\times$ 2& \textbf{46.1} & 83.05 \\
         VADTree & 0.3h & 0.6h &20.0h $\times$ 2& - & 10.7h $\times$ 2  & 62.3 & 84.74 \\
         VADTree-Coarse & 0.3h & 0.2h &5.2h $\times$ 2& - &0.6h $\times$ 2 (no Think) & \textbf{12.1} & 82.83 \\
         VADTree-Fine & 0.3h & 0.4h &14.8h $\times$ 2& - &1.2h $\times$ 2 (no Think) & \textbf{32.7} & 82.72 \\
         VADTree & 0.3h & 0.6h &20.0h $\times$ 2& - &1.8h $\times$ 2 (no Think) & \textbf{44.5} & 84.65 \\
         VADTree-Coarse & 0.3h & 0.2h &5.2h $\times$ 2& - &0.1h $\times$ 2 (t5gemma-9B-2B) & \textbf{11.1} & 82.21 \\
         VADTree-Fine & 0.3h & 0.4h &14.8h $\times$ 2& - &0.2h $\times$ 2 (t5gemma-9B-2B) & \textbf{30.0} & 82.19 \\
         VADTree & 0.3h & 0.6h &20.0h $\times$ 2& - &0.3h $\times$ 2 (t5gemma-9B-2B) & \textbf{41.5} & 84.00 \\
        \bottomrule
        \end{tabular}
            }
\end{table}%
\subsection{Stability Analysis}
We conduct error analysis of~\methodshort on UCF-Crime dataset and report their mean and variance. The randomness of the experimental results mainly comes from the randomness of the generated content during VLM and LLM inference. As shown in Table~\ref{tab:stability}, the $\delta$ across all configurations is statistically insignificant compared to the performance gaps between different methods (Table~\ref{tab:ucf_results} and Table~\ref{tab:xd_results}). This confirms that the observed performance is robust against experimental randomness.
\begin{table}[h!]
  \centering
    \caption{Stability analysis of \methodshort~with different HGTree configurations.}
    \label{tab:stability}
    \footnotesize
            {%
                \begin{tabular}{c|ccc cc}
                \toprule
                 \textbf{Method} &\textbf{Exp-1} &\textbf{Exp-2} &\textbf{Exp-3} &  \textbf{Mean Results}&$\bm{\delta}$ \\
                \midrule
                \methodshort-Coarse & 82.81& 82.75& 82.92& 82.83 & 0.17\\%
                \methodshort-Fine   & 83.05& 82.86& 83.05& 82.99 & 0.19\\%
                \methodshort        & 84.74& 84.49& 84.73& 84.65 & 0.25\\%
                \bottomrule
                \end{tabular}%
            }
\end{table}%
\subsection{Additional Case Studies and Qualitative Results}

\begin{figure}[!t]
\centering
\includegraphics[width=1\linewidth]{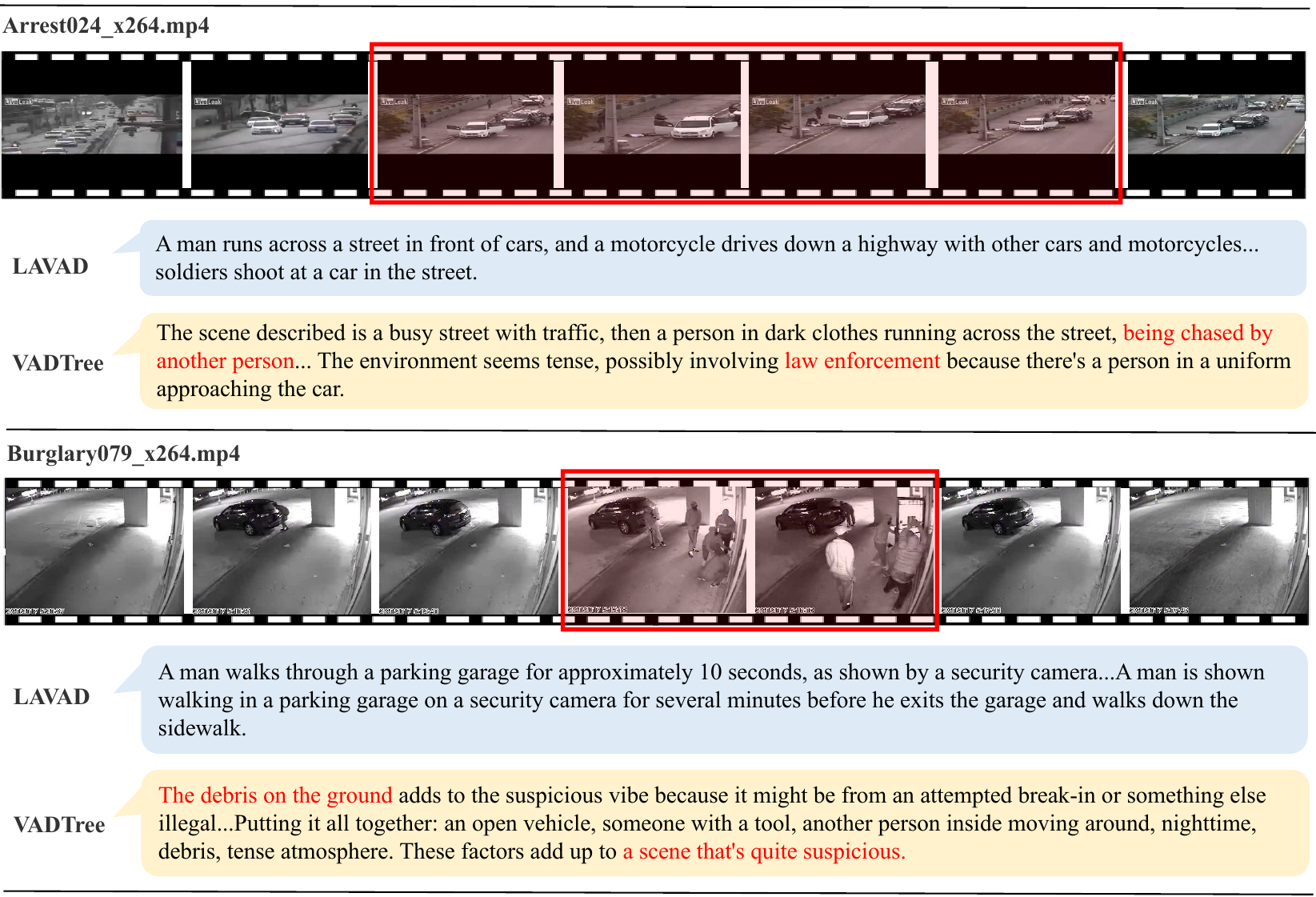}
\caption{ 
Case studies of complex anomalies: arrests and burglaries. VADTree excels in generating accurate explanations by modeling hierarchical events and long-range dependencies, whereas LAVAD produces incomplete descriptions due to frame-level limitations.
}
\label{fig:demo_2imgs}
\end{figure}

For complex anomalies such as arrests and burglaries, the fixed temporal sampling used in LAVAD and frame-level caption aggregation can lead to missed long-range semantic dependencies, resulting in inaccurate or incomplete interpretations of abnormal events. To show the superiority of our VADTree, we conduct a qualitative analysis using two video samples from the UCF-Crime dataset. 
The Figure~\ref{fig:demo_2imgs} displays some key inference information.
LAVAD’s frame-level semantic aggregation often leads to hallucinations and struggles to accurately identify long-range abnormal events. In contrast, VADTree excels at detecting sub-events (e.g., "chased by another person," "attempted break-in") while also synthesizing long-range contextual clues (e.g., "putting it all together"). The above qualitative results will be added to our revision. 

\begin{figure}[!t]
\centering
\includegraphics[width=\linewidth]{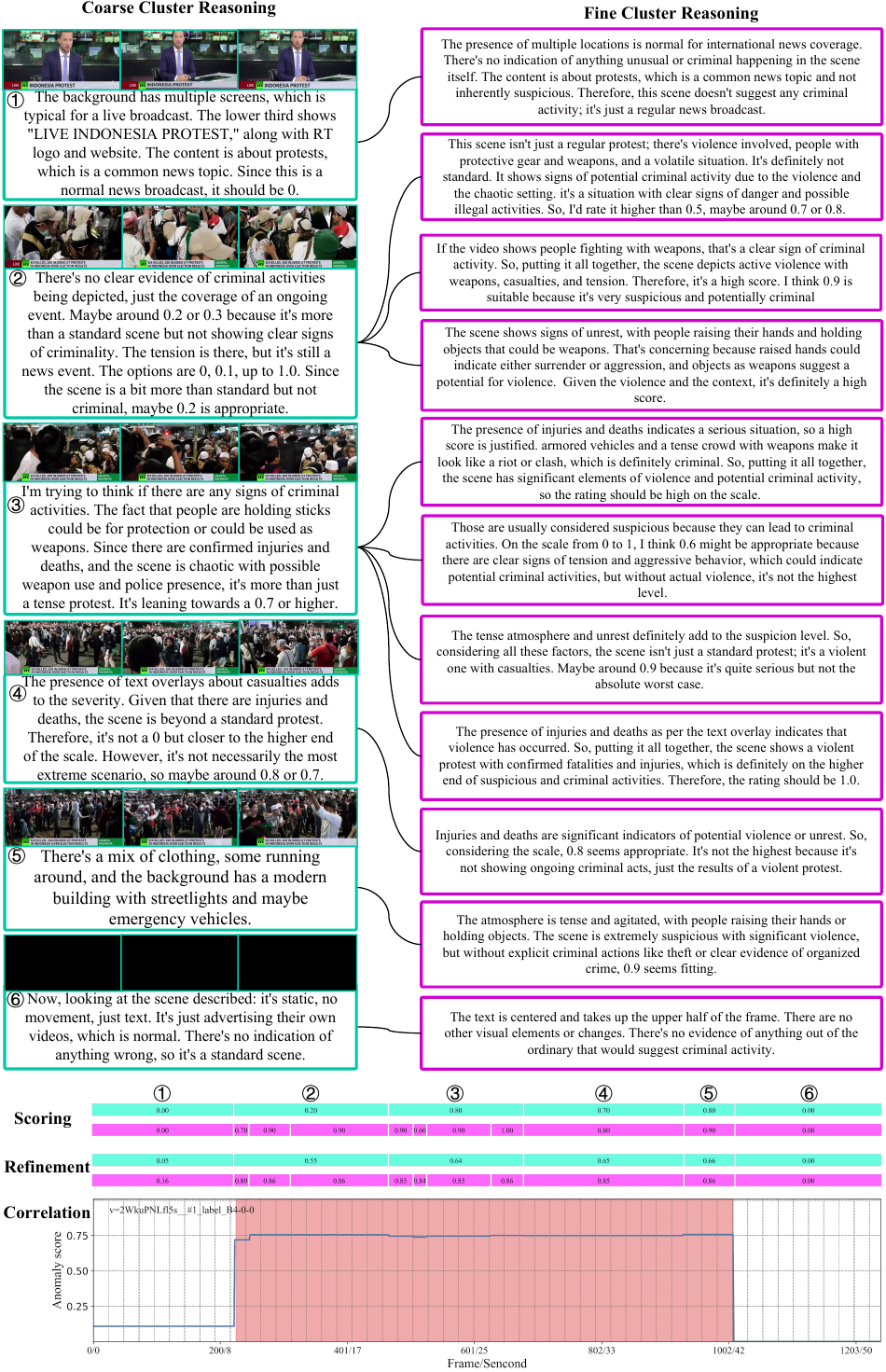}
\caption{Qualitative results of VADTree on a test video, showcasing anomaly explanation and scoring based on HGTree representation. The scoring and text explanation include initial anomaly scores (Scoring), refined scores (Refinement), and final anomaly scores (Correlation) Based on HGTree for video representation, the different granularity reasoning results of coarse and fine clusters on anomalies can complement each other.}
\label{fig:app_demo}
\end{figure}

In the example shown in Figure~\ref{fig:app_demo}, the overall score of fine clusters fluctuates greatly ([0.7, 0.9, 0.9, 0.9, 0.6] in $\Circled{2}$ and [0.9, 0.6, 0.9, 1.0] in $\Circled{3}$), while the anomaly score of coarse cluster $\Circled{2}$ node is low. After refinement, the above situation is improved, but the correct anomaly score is suppressed ($\Circled{3}$: 0.8$\rightarrow$0.64). In addition, the anomaly reasoning of each node is independent and lacks mutual reference and dimensions; this may lead to unstable anomaly scores in the same video segment. The final inter-cluster correlation resolves these shortcomings well. This example indicates that VADTree can significantly enhance the robustness of training-free VAD inference.
\section{Limitations}
Like existing training-free VAD methods, \methodshort's performance relies heavily on the visual perception capabilities of VLMs. 
Most VLMs mainly focus on more complex semantic understanding and reasoning, and there are still significant limitations in accurately obtaining various small shallow abnormal semantics (such as the flame of a lighter after an explosion).
This constraint may hinder accurate anomaly detection. If essential visual characteristics are not captured during the encoding stage, it becomes unlikely for \methodshort to effectively carry out abnormal reasoning or perform temporal inter-cluster corrections.
Therefore, a primary challenge for VLM based VAD is to guarantee that both visual and temporal features are adequately captured.
\section{Broader Societal Impacts}
Our training-free paradigm enables efficient video anomaly detection with minimal computational costs, yet its deployment in safety-critical scenarios (e.g., public surveillance) requires careful consideration of privacy-preserving mechanisms. While avoiding explicit biometric data processing, the prior knowledge base could theoretically retain sensitive environmental patterns. We advocate transparency audits to mitigate potential privacy risks in real-world implementations.
\end{document}